\documentclass[journal]{IEEEtran}

\pdfoutput=1

\ifCLASSINFOpdf
   \usepackage{graphicx}
   \DeclareGraphicsExtensions{.pdf,.jpeg,.png,.jpg}
\else
\fi

\def\compileforpublish{1}

\usepackage[english]{babel}
\usepackage{amsmath}
\usepackage{algorithmic}
\usepackage{stfloats}
\usepackage{url}
\usepackage{tabularx,calc}
\usepackage{color}
\usepackage{tubscolors}

\usepackage{tikz}

\usetikzlibrary{
    shapes,
    arrows,
    angles,
    matrix,
    backgrounds,
    fit,
    patterns,
    decorations.markings,
    shapes.multipart, 
    positioning, 
    calc
}
\usepackage{pgfplots}
\usepackage{pgfplotstable}
\usepackage{tikz-3dplot} 
\usepgfplotslibrary{statistics,groupplots,fillbetween}

\pgfplotsset{compat=1.14}
\tikzset{font={\footnotesize}}

\usepackage[input-protect-tokens, table-number-alignment=center, binary-units=true]{siunitx}
\DeclareSIUnit\LongDegree{deg.}

\usepackage[
    style=ieee,    
    sortlocale=en_US,
    backend=bibtex,
    minnames=1,
    maxcitenames=2,
    maxbibnames=99,
    doi=false,
    isbn=false,
    url=false,
    natbib=true, 
    eprint=false,
    giveninits=true
    ]{biblatex} 
\usepackage[caption=false, font=footnotesize]{subfig}

\usepackage{ifthen}
\usepackage{lipsum}

\usepackage{amssymb} 	
\usepackage{bm} 		
\usepackage{booktabs} 	
\usepackage{multirow}   
\usepackage{colortbl}
\usepackage{enumitem}
\usepackage{lmodern}

\AtBeginDocument{\providecommand\tikzexternaldisable{\relax}} 
\AtBeginDocument{\providecommand\tikzexternalenable{\relax}} 

\newcolumntype{P}[1]{>{\centering\arraybackslash}p{#1}}

\setlist{nosep, leftmargin=1em}

\addto\extrasenglish{}
\addto\extrasenglish{}
\addto\extrasenglish{}
\addto\extrasenglish{}

\usepackage{xpatch}

\xpatchbibdriver{article}
{\usebibmacro{note+pages}}
{\usebibmacro{publisher+location+date}%
    \newunit\newblock
    \usebibmacro{note+pages}}{}{}

\makeatletter
\newcounter{IEEE@bibentries}
\renewcommand\IEEEtriggeratref[1]{%
	\renewbibmacro{finentry}{%
		\stepcounter{IEEE@bibentries}%
		\ifthenelse{\equal{\value{IEEE@bibentries}}{#1}}
		{\finentry\@IEEEtriggercmd}
		{\finentry}%
	}%
}
\makeatother

\ifx \isaccepted \undefined
\newcommand\copyrighttext{%
	\footnotesize \centering This work has been submitted to the IEEE for possible publication.\\ Copyright may be transferred without notice, after which this version may no longer be accessible.}
\else
\newcommand\copyrighttext{%
	\footnotesize \parbox[t]{.11\textwidth}{\copyright{} \the\year~IEEE.} \parbox[t]{.89\textwidth}{Personal use of this material is permitted. Permission from IEEE must be obtained for all other uses, in any current or future media, including reprinting/republishing this material for advertising or promotional purposes, creating new collective works, for resale or redistribution to servers or lists, or reuse of any copyrighted component of this work in other works.}}
\fi
\newcommand\copyrightnotice{%
	\ifx \compileforpublish \undefined
	\else
		\begin{tikzpicture}[remember picture,overlay]
			\node[anchor=south,yshift=22.5pt] at (current page.south) {\parbox{\dimexpr\textwidth-\fboxsep-\fboxrule\relax}{\copyrighttext}};
		\end{tikzpicture}%
	\fi
}

\renewcommand{\vec}[1]{\boldsymbol{#1}}

\providecommand{\function}{}
\renewcommand{\function}[1]{\text{\textsf{#1}}}

\usepackage[left=0.75in,right=0.75in,top=0.75in,bottom=0.75in]{geometry}

\begin{document}
	
\title{A LiDAR-based real-time capable 3D Perception System for Automated Driving in Urban Domains}

\author{
    Jens Rieken %
    and Markus~Maurer
    \thanks{Manuscript received \today}
    \thanks{J. Rieken is a former research assistant at the Institute of Control Engineering, TU Braunschweig, Germany. This journal summarizes the main aspects of his upcoming PhD thesis.}
    \thanks{M. Maurer is professor and head of the Institute of Control Engineering at TU Braunschweig, Germany.}
    }

\maketitle

\newtheorem{definition}{Definition}

\begin{abstract}
    We present a LiDAR-based and real-time capable 3D perception system for automated driving in urban domains.
    The hierarchical system design is able to model stationary and movable parts of the environment simultaneously and under real-time conditions.
    Our approach extends the state of the art by innovative in-detail enhancements for perceiving road users and drivable corridors even in case of non-flat ground surfaces and overhanging or protruding elements.
    We describe a runtime-efficient pointcloud processing pipeline, consisting of adaptive ground surface estimation, 3D clustering and motion classification stages.
    
    Based on the pipeline's output, the stationary environment is represented in a multi-feature mapping and fusion approach. 
    Movable elements are represented in an object tracking system capable of using multiple reference points to account for viewpoint changes.
    We further enhance the tracking system by explicit consideration of occlusion and ambiguity cases.

    Our system is evaluated using a subset of the TUBS Road User Dataset. 
    We enhance common performance metrics by considering application-driven aspects of real-world traffic scenarios.
    The perception system shows impressive results and is able to cope with the addressed scenarios while still preserving real-time capability.
\end{abstract}

\begin{IEEEkeywords}
    Environment perception, LiDAR data processing, Automated Driving
\end{IEEEkeywords}

\copyrightnotice

\xdefinecolor{elevated}{RGB}{0, 128, 255}
\xdefinecolor{ground}{RGB}{0, 0, 0}
\xdefinecolor{curb}{RGB}{0, 196, 128}

\colorlet{common_point}{tuBlueMedium100}

\colorlet{segment_color_1}{tuVioletLight80}
\colorlet{segment_color_2}{tuBlueLight100}
\colorlet{segment_color_3}{tuGreenLight100}
\colorlet{segment_color_4}{tuOrangeLight100}
\colorlet{segment_color_0}{gray!50}

\pgfplotscreateplotcyclelist{tubs-common-4} {%
	tubsBlueLight100\\%
	tubsGreenLight100\\%
	tubsOrangeMedium100\\%
	tubsVioletMedium100\\%
}

\pgfplotsset{
	colormap={tubsBlueDark-CM}{
		color=(tubsBlueDark20);
		color=(tubsBlueDark40);
		color=(tubsBlueDark60);
		color=(tubsBlueDark80);
		color=(tubsBlueDark100);
	},
	colormap={tubsBlueMedium-CM}{
		color=(tubsBlueMedium20);
		color=(tubsBlueMedium40);
		color=(tubsBlueMedium60);
		color=(tubsBlueMedium80);
		color=(tubsBlueMedium100);
	},
	colormap={tubsBlueLight-CM}{
		color=(tubsBlueLight20);
		color=(tubsBlueLight40);
		color=(tubsBlueLight60);
		color=(tubsBlueLight80);
		color=(tubsBlueLight100);
	},
	colormap={tubsGreenDark-CM}{
		color=(tubsGreenDark20);
		color=(tubsGreenDark40);
		color=(tubsGreenDark60);
		color=(tubsGreenDark80);
		color=(tubsGreenDark100);
	},
	colormap={tubsGreenMedium-CM}{
		color=(tubsGreenMedium20);
		color=(tubsGreenMedium40);
		color=(tubsGreenMedium60);
		color=(tubsGreenMedium80);
		color=(tubsGreenMedium100);
	},
	colormap={tubsGreenLight-CM}{
		color=(tubsGreenLight20);
		color=(tubsGreenLight40);
		color=(tubsGreenLight60);
		color=(tubsGreenLight80);
		color=(tubsGreenLight100);
	},
	colormap={tubsOrangeDark-CM}{
		color=(tubsOrangeDark20);
		color=(tubsOrangeDark40);
		color=(tubsOrangeDark60);
		color=(tubsOrangeDark80);
		color=(tubsOrangeDark100);
	},	
	colormap={tubsOrangeMedium-CM}{
		color=(tubsOrangeMedium20);
		color=(tubsOrangeMedium40);
		color=(tubsOrangeMedium60);
		color=(tubsOrangeMedium80);
		color=(tubsOrangeMedium100);
	},
	colormap={tubsOrangeLight-CM}{
		color=(tubsOrangeLight20);
		color=(tubsOrangeLight40);
		color=(tubsOrangeLight60);
		color=(tubsOrangeLight80);
		color=(tubsOrangeLight100);
	},	
	colormap={tubsVioletMedium-CM}{
		color=(tubsVioletMedium20);
		color=(tubsVioletMedium40);
		color=(tubsVioletMedium60);
		color=(tubsVioletMedium80);
		color=(tubsVioletMedium100);
	},
}
\pgfplotsset{
	/pgfplots/colormap={jet}{rgb255(0cm)=(0,0,128) rgb255(1cm)=(0,0,255)
		rgb255(3cm)=(0,255,255) rgb255(5cm)=(255,255,0) rgb255(7cm)=(255,0,0) rgb255(8cm)=(128,0,0)}
}

\section{Introduction}

Environment perception is a crucial component for automated driving functions.
The perception provides input for driving decision tasks and vehicle control strategies.
As a consequence, its performance defines the extent to which automated driving functions can be realized.
 
 Perception systems have to be able to detect and represent various types of traffic participants in partly unexpected surroundings.
This holds true especially for urban domains with its high traffic density and complex lane and road structures. 
Combined with a variety of different types of peripheral infrastructure and vegetation, these aspects make perception a challenging task.
Protruding loads, overhanging vegetation or non-flat road surfaces need to be dealt with appropriately, thus it is mandatory to perceive the environment in 3D.
Next to other sensor technologies, much effort has been spent on LiDAR-based perception systems throughout the last decade.
Especially \emph{high-resolution} LiDAR sensors like Velodyne's HDL-64E series 
are enablers for 3D perception due to their large vertical field of view.
However, the amount of data generated by such a sensor system is computationally demanding, especially for real-world and closed-loop driving applications.

While many published research activities focus on specific aspects of the perception system and on enhancing such algorithms, the ability to run the entire perception system as part of an experimental vehicle under real-time constraints has gotten comparably little attention.
In addition, only few publications address full-scale, high-resolution LiDAR perception systems that are able to represent the stationary as well as the movable environment simultaneously.  

\subsection{Contribution of this paper}
In this paper, we present the high-resolution LiDAR-based \ang{360} perception system of the project \emph{Stadtpilot}, which addresses automated driving in the inner-city domain \citep{Wille_Stadtpilot_2010, Nothdurft_Stadtpilot_2011}. 
We describe a hierarchical system design that is able to represent stationary and movable elements in 3D.
In conjunction with an efficient pointcloud representation, it is shown that the system is capable of real-time execution on vehicle-mounted hardware and, by that, is applicable for closed-loop driving functions.
This paper addresses the following aspects in deeper detail:
\begin{itemize}
    \item Usage of an adaptive ground surface estimation, which serves as the basis for a surface-relative environment representation,
    \item a multi-stage pointcloud preprocessing pipeline, including pointcloud compression and clustering,
    \item a motion classification of detected clusters, and
    \item a hybrid representation for movable and stationary environment elements, consisting of a multi-layer semantic grid map and an IMM-EKF-based object tracking with explicit modeling of partial occlusions.  
\end{itemize}
We present the latest additions to our previous work (e.\,g.  \citep{Rieken_Environment_Modelling_2015, Matthaei_Grid_Road_Detection_2013, Matthaei_Lane_Level_Localization_2014, Matthaei_Consistency_FUSION_2011, Rieken_Ground_Surface_2015, Matthaei_Grids_2014, Rieken_Scan_Timing_2016, Choi_SLAM_Tracking_2016}) 
and provide evaluation results of the perception performance using a subset of the TUBS Road User Dataset~\citep{Plachetka_TUBS_Dataset_2018}. 
We extend common evaluation strategies by introducing application- and scenario-specific relevance criteria to the evaluation process.

Note that, due to length restrictions, we cannot provide in-depth algorithmic discussions, but rather an architectural overview of the perception system only. 
The reader is referred to the cited project-related publications for further details.
Also note that localization-related topics are excluded in this paper.
Throughout the remaining parts we assume that the vehicle's pose within a locally fixed coordinate frame is available.

\providecommand{\currentimagedirectory}{}
\renewcommand{\currentimagedirectory}{./figures/3_System_Design}

\begin{figure*}
    \centering
    \begin{tikzpicture}[align=center, node distance = 0.45cm and 1cm]

\pgfdeclarelayer{background}
\pgfsetlayers{background,main}

 \makeatletter
  \tikzset{minimum dist/.style 2 args={%
    insert path={%
      \pgfextra{%
       \path (#1);
       \pgfgetlastxy{\xa}{\ya} 
        \path (#2);
       \pgfgetlastxy{\xb}{\yb}   
       \pgfpointdiff{\pgfpoint{\xa}{\ya}}%
                     {\pgfpoint{\xb}{\yb}}%
       \pgf@xa=\pgf@x}
        },
    minimum width=\pgf@xa}
    } 
    \makeatother
 
\input{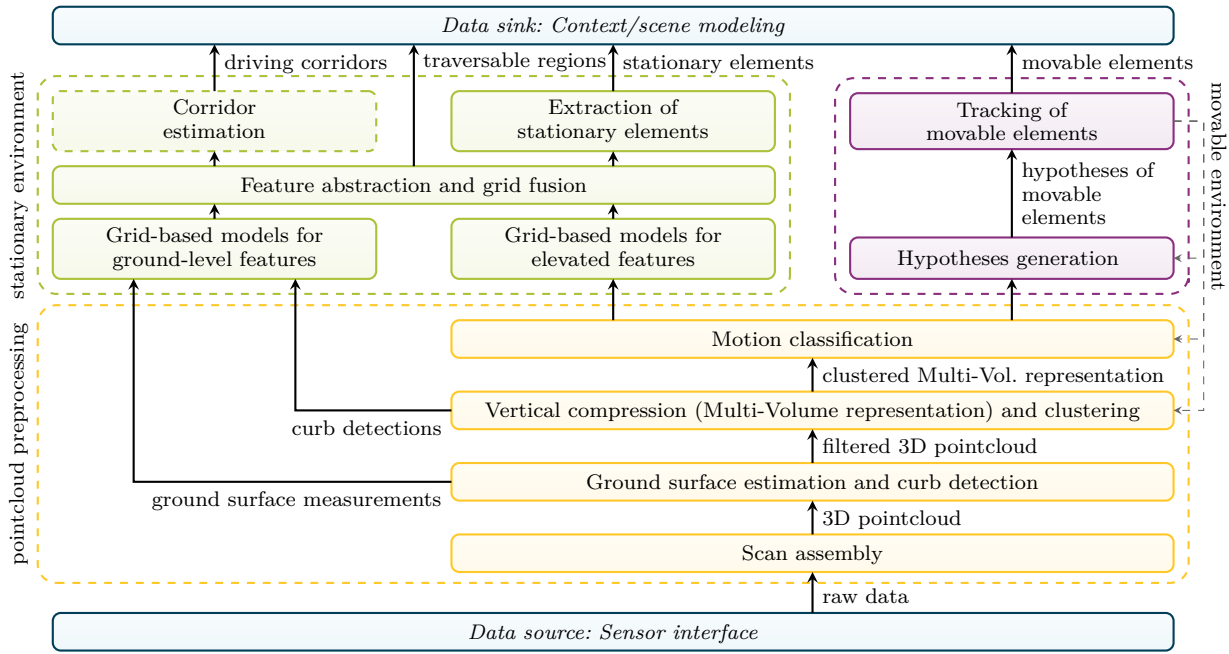}

\tikzstyle{componenttext}=[anchor=mid, font={\footnotesize}, yshift=-0.1ex];		
	
\tikzstyle{Feedback}= [Transition, dashed,thin, tubsGray60];
	
\tikzstyle{domainborder} = [black!50, dashed]

\node (HypothesesGeneration) [MovableComponent,
	]{ \\ };
\draw (HypothesesGeneration) node [componenttext]	{Hypotheses generation };

\node (GridModels) [StationaryComponent,
	left = of HypothesesGeneration.south west,
	anchor = south east,
	] {Grid-based models for \\ elevated features\vphantom{y}};

\node (3DGridModels) [StationaryComponent,
	left = of GridModels,
	] {Grid-based models for \\ ground-level features\vphantom{y}};

\node (GridFusion) [StationaryComponent,
	above = of GridModels.north east,
	yshift=-0.25cm,
	anchor = south east,
	minimum dist={3DGridModels.west}{GridModels.east}
	] {};
\draw(GridFusion) node [componenttext]	{Feature abstraction and grid fusion};
	
\node (BlockageDetection) [StationaryComponent,
	above = of GridFusion.north east,
	anchor = south east,
		yshift=-0.25cm
	]
	{Extraction of\\stationary elements};
	
\node (RoadExtraction) [StationaryComponent, dashed,
	left = of BlockageDetection.north west,
	anchor = north east,
	minimum dist={3DGridModels.west}{3DGridModels.east}
	]
	{Corridor\\estimation\vphantom{y}};

\node (ObjectTracking) [MovableComponent,
	above = of HypothesesGeneration,
	right = of BlockageDetection,
	] {Tracking of \\ movable elements};
	
\node (DynamicClassification) [PreprocessingComponent,
	below = of HypothesesGeneration.south east,
	yshift=-0.1cm,
	anchor = north east,
	minimum dist={GridModels.west}{HypothesesGeneration.east}]
	{};
\draw (DynamicClassification.center) node[componenttext] {Motion classification};

\node (Segmentation)[PreprocessingComponent,
	below = of DynamicClassification,
	minimum dist={DynamicClassification.west}{DynamicClassification.east},
	] {};
\draw (Segmentation.center) node[componenttext] {Vertical compression (Multi-Volume representation) and clustering};

\node (3DFeatureExtraction)[PreprocessingComponent,
	below = of Segmentation,
	minimum dist={Segmentation.west}{Segmentation.east}
	] {};
\draw(3DFeatureExtraction) node [componenttext]	{Ground surface estimation and curb detection};

\node (ContextModel) [OutOfScopeComponent,
	above = of ObjectTracking.north east,
	anchor = south east,
	minimum dist = {RoadExtraction.west}{ObjectTracking.east},
	yshift=0.2cm
	] {};
\draw(ContextModel) node [componenttext, dashed]	{\emph{Data sink: Context/scene modeling}};

\node (SensorDataCollection)[PreprocessingComponent,
	below = of 3DFeatureExtraction.south east,
	anchor = north east,
	minimum dist = {3DFeatureExtraction.west}{3DFeatureExtraction.east},
	]	{};
\draw (SensorDataCollection) node [componenttext] {Scan assembly};

\node (SensordataAcquisition)[OutOfScopeComponent,
	below = of SensorDataCollection.south east,
	anchor = north east,
	yshift=-0.095cm,
	minimum dist = {ContextModel.west}{ContextModel.east},
	]	{};
	
\draw(SensordataAcquisition) node [componenttext] {\emph{Data source: Sensor interface}};
	
\draw [Transition]		  (Segmentation)          -- (DynamicClassification) node[midway, anchor=north west, yshift=0.65em] 
	{clustered Multi-Vol. representation}; 	
	
\draw [ReverseTransition] 		  (3DFeatureExtraction.south) -- (SensorDataCollection.north) node [midway, right, anchor=north west, yshift=0.65em]
	{3D pointcloud};
	
\draw [ReverseTransition] 		  (SensorDataCollection.south) -- (SensorDataCollection.south |- SensordataAcquisition.north) node [at end, right, anchor=south west]
	{raw data};
	
\draw [ReverseTransition] (Segmentation)          -- (Segmentation|-3DFeatureExtraction.north) node [midway, right, anchor=north west, yshift=0.65em]
	{filtered 3D pointcloud};

\draw [ReverseTransition] (HypothesesGeneration)  -- (HypothesesGeneration|-DynamicClassification.north);
	
\draw [ReverseTransition] (GridModels)            -- (GridModels|-DynamicClassification.north);

\coordinate (NonMovableSegmentsHorizontalCenter) at ($(GridModels.south)!0.5!(HypothesesGeneration.south)$);
\coordinate (NonMovableSegmentsVerticalBottom) at (GridModels.south |- DynamicClassification.north);


\draw [Transition]		  (HypothesesGeneration)  -- (ObjectTracking) node [midway, right, align=left]  {hypotheses of \\ movable \\ elements};
\draw [Transition]		  (3DFeatureExtraction)   -|  ($(3DGridModels.south west)!0.5!(3DGridModels.south)$) node [at start, below, anchor=north east] {ground surface measurements};
\draw[Transition]		(Segmentation) -| ($(3DGridModels.south)!0.5!(3DGridModels.south east)$) node [at start, below, align=left, anchor=north east]  {curb detections};
\draw [Transition]		  (3DGridModels)		  -- (GridFusion.south-|3DGridModels);
\draw [Transition]		  (GridModels)		  	  -- (GridFusion.south-|GridModels);
\draw [ReverseTransition] (BlockageDetection)     -- (BlockageDetection|-GridFusion.north);
\draw [ReverseTransition] (RoadExtraction)        -- (RoadExtraction|-GridFusion.north);
\draw [Transition]	      (RoadExtraction)        -- (RoadExtraction|-ContextModel.south) node [at end, right, align=center,anchor= north west] {driving corridors};
\draw [Transition]	      (ObjectTracking)        -- (ObjectTracking|-ContextModel.south) node [at end, right, align=center, anchor=north west] (contextmodelinputcaption){movable elements};
\draw [Transition]	      (BlockageDetection)     -- (BlockageDetection|-ContextModel.south) node [at end, right, align=center, anchor=north west] {stationary elements};
\draw [Transition] (GridFusion) -- (ContextModel.south-|GridFusion);
\node[right] at (contextmodelinputcaption -| GridFusion) {traversable regions};

\draw[Feedback] (ObjectTracking.east) -- ++ (0.40,0) |- (Segmentation.east);
\draw[Feedback] (ObjectTracking.east) -- ++ (0.40,0) |- (DynamicClassification.east);
\draw[Feedback] (ObjectTracking.east) -- ++ (0.40,0) |- (HypothesesGeneration.east);

\coordinate (lowerdomainvertical) at ($(SensordataAcquisition.north)+(0,0.15)$);
\coordinate (upperdomainvertical) at ($(ContextModel.south) + (0,-0.15)$);

\tikzstyle{chapterborder}=[dashed, rounded corners=0.25cm, tubsGray, thick, font={\footnotesize}];
\begin{pgfonlayer}{background}
	\draw [chapterborder, preprocessingcolor]
				   ($(SensordataAcquisition.south west) + (-0.20, 0.9)$) 
				-- ($(SensordataAcquisition.south east) + (0.20,0.9)$)
				-- ($(DynamicClassification.north east) + (0.20, 0.20)$)
				-- ($(SensordataAcquisition.south west |- DynamicClassification.north east)  + (-0.20,0.20)$) 
				-- cycle node[midway, above, sloped, rotate=180, black] {pointcloud preprocessing};
			
	\draw[chapterborder, movablecolor]
				($(HypothesesGeneration.south east) + (0.2,-0.2)$)
			--	($(ObjectTracking.north east) + (0.2, 0.2)$)
				node [midway, sloped, above, rotate=180, black, yshift=0.5em] {movable environment}
			--  ($(ObjectTracking.north west) + (-0.2, 0.2)$) 
			|- cycle;
	\draw[chapterborder, stationarycolor]
				($(GridModels.south east) + (0.2, -0.2)$)
			--	($(BlockageDetection.north east) + (0.2, 0.2)$)
			--  ($(RoadExtraction.north west) + (-0.2, 0.2)$)
			--  ($(3DGridModels.south west) + (-0.2,-0.2)$)
				node[midway, above, sloped, rotate=180, black] {stationary environment}
			--  cycle ;
\end{pgfonlayer}
\end{tikzpicture}
    \vspace*{-0.5em}
    \caption{Refined dataflow architecture of the presented 3D perception system, based on first results of \citet{Rieken_Environment_Modelling_2015}. Data processing consists of three major blocks, starting with the \emph{pointcloud preprocessing}. A motion classification stage then assigns the refined pointcloud to the \emph{stationary environment} and \emph{movable environment} modeling. Both models provide their results to the context modeling module. Although being part of this architecture, the \emph{Corridor estimation} is not addressed within this paper.}
    \label{fig:Sytem_Design_Architecture}
    \vspace{-1em}
\end{figure*}

\subsection{Paper outline}
This paper is organized as follows.
We summarize the current state of the art regarding full-scale perception systems in \autoref{sec:Related_work}.
We continue with a description of the main aspects of our perception system.
Its dataflow architecture is shown in \autoref{fig:Sytem_Design_Architecture}.
It represents one possible realization of the \emph{feature-extraction and model-based filtering} component of the functional system architecture by \citet{Matthaei_FSA_2015}.
Aspects of efficient pointcloud management and preprocessing stages are explained in \autoref{sec:Preprocessing}.
The preprocessing modules separate ground measurements from elevated targets and provide cluster information.
Based on the results of a motion classification, clusters are either forwarded to the stationary or the movable environment model (\autoref{sec:Stationary_Environment_Modeling} resp. \autoref{sec:Object_Tracking}).
The former model represents the stationary parts of the environment, such as elevated elements close to or on the road, information about traversable regions, and driving corridors.
The latter model is dedicated to movable elements, i.\,e., other traffic participants. 
Results are provided to the project's context modeling module, which serves as a data sink for the perception system. 
We evaluate the results of the perception system in \autoref{sec:Evaluation}.
and conclude this paper with a summary in \autoref{sec:Conclusion}.

\section{Overview of similar work}
\label{sec:Related_work}

Since the DARPA Urban Challenge in \num{2007}, high-resolution LiDAR systems are chosen for environment perception tasks in larger scales.
Although many publications address specific aspects of relevant processing stages, only few groups seem to target LiDAR-based perception systems with simultaneous representation of stationary and movable environments.
In the following paragraphs, we will provide a short review of published perception systems that use such sensors as the primary source of information.

Beginning in \num{2008}, \citet{Himmelsbach_Perception_2008, Himmelsbach_Segmentation_2010} published approaches for ground point classification, grid-based clustering and object detection.
Bounding-box hypotheses are  tracked over time \citep{Himmelsbach_Tracking_Detection_2012}.
\citet{Manz_Road_Detection_2011} and \citet{Jaspers_Map_generation_2017} use elevation mapping techniques for road and ground surface estimation in rural environments. 
\citeauthor{Jaspers_Map_generation_2017} also include information from the tracking system of \citet{Himmelsbach_Tracking_Detection_2012} to exclude moving parts from the mapping process.
Recent publications focus on enhancing clustering and hypotheses generation algorithms \citep{Naujoks_Bounding_Boxes_2018, Burger_Segmentierung_2018}.
As one of few groups which regularly participate in land robot trials, their work considers real-time constraints and has been shown to work in closed-loop driving.
So far, their approaches do not regard issues present in urban domains, such as occlusions or viewpoint changes.

\citet{Moosmann_Segmentation_2009} published a perception system based on \emph{local convexity criteria}.
They treat scan data as depth images and calculate local range gradients, which then serve as features for clustering, motion detection  and SLAM algorithms \citep{Moosmann_Segmentation_2009, Moosmann_Motion_Estimation_2010, Moosmann_SLAM_2011}, as well as for movable object tracking \citep{Moosmann_Dissertation_2013}.
Their tracker is able to handle arbitrary object shapes.
In contrast to others, they do not apply different modeling techniques for different filter steps, but use local convexity features as a unified representation.
Despite promising results, they stated their approach as not being real-time capable \citep{Moosmann_Dissertation_2013}.

\citet{Petrovskaya_Vehicle_Detection_Tracking_2009} published a particle-filter-based approach for object tracking that was also applied in the DARPA Urban Challenge.
\citet{Levinson_Dissertation_2011} provided mapping and calibration approaches based on LiDAR measurements.
In later work, \citet{Teichman_Tracking_2011, Teichman_Semi_Supervised_Learning_2012} presented approaches for 3D object classification and tracking. 
\citet{Held_Tracking_2013, Held_Robust_Tracking_2014} extended this work by accumulating shape information along with color data from camera images. 
The same modalities, along with knowledge from previous time steps, are introduced by \citet{Held_Segmentation_2016} to enhance the clustering process. 
In general, their tracking and clustering algorithms were stated as real-time capable relative to the sensor's update rate.
Yet, results neither for large numbers of simultaneously appearing objects nor the performance in real-world driving applications have been published.

\citet{Asvadi_Detection_2016} presented a perception system that uses a multi-region plane approach for ground surface detection and a voxel representation for non-ground elements.
Although their approach is able to address the aforementioned challenges of 3D perception, they rated their system as not being able to run in real-time.
\citet{Azim_Moving_Tracking_2012} presented an approach for SLAM and object tracking using an octree data structure.
Unfortunately, no information regarding execution time was published.
\citet{Choi_SLAM_Tracking_2016} use a similar data structure.
They combine feature- and map-based SLAM algorithms with a bounding-box tracker and provide solutions for handling object shape changes \citep{Choi_Tracking_2013}.
While their pointcloud preprocessing stages and object tracking system are real-time capable, the mapping part exceeds the runtime limitations.
In addition, their preprocessing stages are not able to handle vertically stacked elements like vehicles beneath trees.

Despite impressive results, none of the aforementioned approaches have published quantitative evaluations regarding their real-world performance for common driving tasks in urban environment, such as ACC following and lane changing.
Only very  recent  LiDAR-based object tracking systems (e.\,g. \citep{Kampker_Multi_Object_Detection_2018, Sualeh_Velodyne_Tracking_2019}) provide results in terms of statistical values and tracking-specific metrics, acquired from a common dataset such as KITTI \citep{Geiger_KITTI_2012}.
Even though the commonly applied metrics allow a comparison of different approaches, they lack a possibility to rank between scenario-specific errors.
This makes their meaningfulness for the evaluation of the perception's performance for actual driving tasks questionable.
Further details about this topic will be given in Sec.~\ref{sec:Evaluation}.

\section{Runtime efficient Pointcloud preprocessing}
\label{sec:Preprocessing}
\providecommand{\currentimagedirectory}{}
\renewcommand{\currentimagedirectory}{./figures/4_Preprocessing}
Due to the large amount of measurement data, processing stages that deal with pointcloud data must be designed to be runtime efficient.
Next to algorithmic decisions, an efficient data storage and access concept is a key factor to achieve this goal.
In the following, we first present a pointcloud management scheme  for this purpose.
Next, we explain the main components of our pointcloud preprocessing pipeline, as shown in \autoref{fig:Sytem_Design_Architecture}.

\subsection{Pointcloud management and operation principles}
\label{ssec:Preprocessing_Dataformat}

\begin{figure}
    \centering
    \tikzexternalenable
    \tdplotsetmaincoords{45}{-25}
\begin{tikzpicture}[tdplot_main_coords]

\colorlet{fovcolor}{gray}

\tikzstyle{arrow}=[-stealth'];
\tikzstyle{coordinateaxis}=[arrow, solid];
\tikzstyle{sphericalaxis}=[coordinateaxis, solid];
\tikzstyle{fieldofview}=[fovcolor, densely dashed, thin];
\tikzstyle{containergrid}=[black!35, very thin];
\tikzstyle{containergridborder}=[containergrid, tubsBlueMedium80];

\pgfmathsetmacro{\FOVVBegin}{-5};
\pgfmathsetmacro{\FOVVRange}{10};
\pgfmathsetmacro{\FOVVEnd}{\FOVVBegin + \FOVVRange};
\pgfmathsetmacro{\FOVHBegin}{-15};
\pgfmathsetmacro{\FOVHRange}{30};
\pgfmathsetmacro{\FOVHEnd}{\FOVHBegin + \FOVHRange};

\pgfmathsetmacro{\FOVVStepSize}{2}
\pgfmathsetmacro{\FOVVBeginPlusStep}{\FOVVBegin + \FOVVStepSize}
\pgfmathsetmacro{\FOVVBeginPlusTwoStep}{\FOVVBegin + 2*\FOVVStepSize}
\pgfmathsetmacro{\FOVVEndMinusStep}{\FOVVEnd - \FOVVStepSize}
\pgfmathsetmacro{\FOVHStepSize}{1}
\pgfmathsetmacro{\FOVHBeginPlusStep}{\FOVHBegin + \FOVHStepSize}
\pgfmathsetmacro{\FOVHBeginPlusTwoStep}{\FOVHBegin + 2*\FOVHStepSize}
\pgfmathsetmacro{\FOVHEndMinusStep}{\FOVHEnd - \FOVHStepSize}
\pgfmathsetmacro{\GridRange}{5.5}
\pgfmathsetmacro{\GridRangeSecondary}{7}

\pgfmathsetmacro{\TargetPhi}{4.5}
\pgfmathsetmacro{\TargetTheta}{4}
\pgfmathsetmacro{\TargetRange}{7.5}
\pgfmathsetmacro{\TargetRangeSecond}{8.5}

\pgfmathsetmacro{\TargetPhiGrid}{4}
\pgfmathsetmacro{\TargetThetaGrid}{3}
\tikzset{cs/3d/.code args={#1:#2:#3}{\pgfpointxyz{(#1)*cos(#2)*cos(#3)}{(#1)*cos(#2)*sin(#3)}{(#1)*sin(#2)}}};
\tikzdeclarecoordinatesystem{3d}{\tikzset{cs/3d={#1}}}%
 
\tikzset{spherical patch/.style args={#1:#2:#3:#4:#5}{insert path={
  [smooth, samples=50, line join=round]
  plot [domain=0:#3] (3d cs:{#5}:{#1+\x}:{#2}) --
  plot [domain=0:#4] (3d cs:{#5}:{#1+#3}:{#2+\x}) --
  plot [domain=#3:0] (3d cs:{#5}:{#1+\x}:{#2+#4}) --
  plot [domain=#4:0] (3d cs:{#5}:{#1}:{#2+\x}) -- cycle}}}
 
\tikzset{horizontal line/.style args={#1:#2:#3:#4}{insert path={
  [smooth, samples=50, line join=round]
  plot [domain=0:#3] (3d cs:{#4}:{#1}:{#2+\x})}}}
 
\tikzset{vertical line/.style args={#1:#2:#3:#4}{insert path={
  [smooth, samples=50, line join=round]
  plot [domain=0:#3] (3d cs:{#4}:{#1+\x}:{#2})}}}

\foreach \j in {\FOVVBeginPlusStep,\FOVVBeginPlusTwoStep,...,\FOVVEndMinusStep}
{
	\draw [containergrid, horizontal line={\j:\FOVHBegin:\FOVHRange:\GridRange}];
	\draw [containergrid, horizontal line={\j:\FOVHBegin:\FOVHRange:\GridRangeSecondary}];
}

\foreach \j in {\FOVHBeginPlusStep,\FOVHBeginPlusTwoStep, ...,\FOVHEndMinusStep}
{
    \draw [containergrid, vertical line={\FOVVBegin:\j:\FOVVRange:\GridRange}];
    \draw [containergrid, vertical line={\FOVVBegin:\j:\FOVVRange:\GridRangeSecondary}];
}
    
\draw [containergridborder, spherical patch={\FOVVBegin:\FOVHBegin:\FOVVRange:\FOVHRange:\GridRange}];
\draw [containergridborder, tubsOrangeMedium80, spherical patch={\FOVVBegin:\FOVHBegin:\FOVVRange:\FOVHRange:\GridRangeSecondary}]; 

\foreach \j/\k in {\FOVVBegin/\FOVHBegin, \FOVVEnd/\FOVHBegin, \FOVVEnd/\FOVHEnd, \FOVVBegin/\FOVHEnd}
{
	\draw[fieldofview] (3d cs:0:\j:\k) -- (3d cs:\GridRangeSecondary:\j:\k);
}
 
\draw[coordinateaxis] (3d cs:\GridRange:\FOVVBegin:\FOVHBegin) -- (3d cs:\GridRange:\FOVVBegin:\FOVHBegin+7) node[at end, left, anchor=north east] {\text{channel}};   
\draw[coordinateaxis] (3d cs:\GridRange:\FOVVBegin:\FOVHBegin) -- (3d cs:\GridRange:\FOVVBegin+7:\FOVHBegin) node[right] {\text{layer}}; 
\draw[coordinateaxis] (3d cs:\GridRange:\FOVVBegin:\FOVHBegin) -- (3d cs:\GridRange+0.6:\FOVVBegin:\FOVHBegin) node[at end, below, anchor=north] {\text{target/return index}}; 

\coordinate (Target) 				at (3d cs:\TargetRange:\TargetTheta:\TargetPhi);
\coordinate (TargetSecond) 				at (3d cs:\TargetRangeSecond:\TargetTheta:\TargetPhi);
\coordinate (TargetAtGrid) 			at 	(3d cs:\GridRange:\TargetTheta:\TargetPhi);
\coordinate (TargetAtGridSecond) 	at (3d cs:\GridRangeSecondary:\TargetTheta:\TargetPhi);

\draw[arrow, black, thick] (TargetAtGridSecond) -- (Target) node [right, anchor =north west] {$r_\text{T0}$};
\draw[arrow, black, thick] (Target) -- (TargetSecond) node [right, anchor=north west] {$r_\text{T1}$};

\draw[tubsBlack, fill=tubsBlueMedium100] (Target) circle (2pt);
\draw[tubsBlack, fill=tubsOrangeMedium100] (TargetSecond) circle (2pt);

\draw[tubsOrangeMedium100, fill=tubsOrangeMedium60, spherical patch={\TargetThetaGrid:\TargetPhiGrid:\FOVVStepSize:\FOVHStepSize:\GridRangeSecondary}];
\draw[black, thick] (TargetAtGrid) -- (TargetAtGridSecond);

\draw[tubsBlueMedium100, fill=tubsBlueMedium60, spherical patch={\TargetThetaGrid:\TargetPhiGrid:\FOVVStepSize:\FOVHStepSize:\GridRange}];
\draw[black, thick] (0,0,0) -- (TargetAtGrid);

\draw[gray, dashdotted] (0,0,0) -- (3d cs:5.1:0:\TargetPhi);
\draw (3d cs:4:0:\TargetPhi) node [gray, right, below, anchor=north west] {$r_\text{T, Zyl.}$};

\draw[coordinateaxis, tubsGreenMedium100]  (0,0,0) -- (0,0.75,0) node[left]  {$y$};
\draw[coordinateaxis, tubsBlueMedium100]  (0,0,0) -- (3.3,0,0) node[midway, below] {$x$};
\draw[coordinateaxis, tubsRed]  (0,0,0) -- (0,0,1) node[above] {$z$};

\draw[sphericalaxis, vertical line={0:\TargetPhi:\TargetTheta:5}];
\draw (3d cs:5.2:-7:\TargetPhi) node {$\theta_\text{T}$};

\draw[sphericalaxis, horizontal line={0:0:\TargetPhi:3}];
\draw (3d cs:3.3:0:-1) node [below] {$\varphi_\text{T}$};
\end{tikzpicture}
    \tikzexternaldisable
    \vspace*{-2em}
    \caption{Representation of the laser scan as a discrete-angle spherical 2D grid with multiple (here: two) slices, depending on the number of supported returns per firing of the sensor. The cells of each slice are shown at the same radial distance for visual purposes only. Each cell stores the respective target's distance ($r_{T0}$, $r_{T1}$). }
    \label{fig:ISDContainer}
    \vspace{-1.5em}
\end{figure}
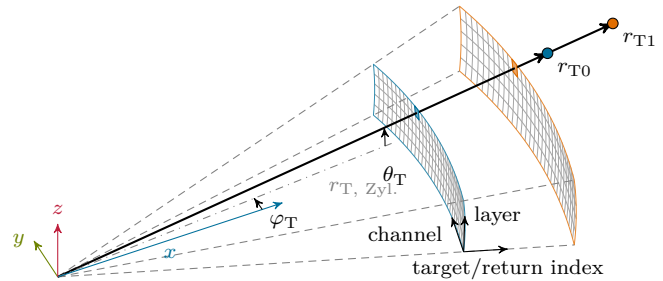 

Sensor data is represented by a spherical grid structure, similar to a depth image.
This structure implicitly defines neighborhood relations and thus allows access to neighboring points in constant time.
To avoid additional discretization, azimuth and elevation angle step sizes are derived from the sensor's characteristics. 
The underlying principle is illustrated in \autoref{fig:ISDContainer}.

Each measurement is assigned to a logical position within the grid.
Angular coordinates are translated into \emph{channel} and \emph{layer} information.
For a single-target LiDAR sensor, i.\,e., a device that provides only a single range measurement per firing, a single \emph{slice} is sufficient to store the measurement information.
For multi-target sensors, additional slices are incorporated; a target index is applied to access the additional return data.
Due to the grid structure, information about \emph{unavailable} measurements are preserved, as some cells of the grid might not be populated, but still constitute valid logical positions.

Throughout the data processing stages, each logical position is augmented by a number of \emph{attributes}, such as cluster labels, but also a set of boolean flags.
The key principle is that the \emph{order} and \emph{measured states} of the measurements are never altered. 
Instead, the algorithms only modify the attributes of the measurements. 
The boolean flags store arbitrary binary states. 
They also serve as flow control within the algorithms.
By this design, it is ensured that information about each measurement's neighborhood remains intact even when filters are applied to the pointcloud.

\subsection{Scan definition and timing behavior}
Measurements from one full turn of the sensor are collected and processed as one single pointcloud. 
This collection is referred to as \emph{scan}\footnote{Another commonly used term is \emph{frame}, borrowed from the computer vision community. 
    The term \emph{frame} is not used here on purpose, because it connotes that all data is captured at a single time instant.}.
Typically, rotating sensor devices do not gather measurements in a snapshot manner, but in an iterative working principle.
Thus, (almost) every measurement is captured at a different time instant.
This fact has to be considered if motion of the sensor platform and/or the perceived targets is present.
In this case, the incremental changes of the relative positions of sensor and targets result in distortions, known as \emph{motion scan effect} \citep{Groll_Motion_Scan_2007}.

In order to tackle the implications of this effect, the measurement timestamp of each single point has to be considered during the process.
Unlike other approaches, which compensate for ego motion by transforming all measurements to a given point in time (e.\,g. \citep{Schneider_LIDAR_Camera_Fusion_2010, Varga_Perception_Semantic_Pointcloud_Labeling_2017, Himmelsbach_Perception_2008, Moosmann_Segmentation_2009}), we propagate the time information through the preprocessing steps.
In contrast to the references mentioned above, this allows a proper consideration of ego \emph{and} target motion, as we have discussed in \citet{Rieken_Scan_Timing_2016}.

	\begin{figure*}
    \centering
    \subfloat[Ground surface and curb estimation]{\label{subfig:Preprocessing_Example_Ground_Surface}%
        \includegraphics[width=0.32\textwidth]{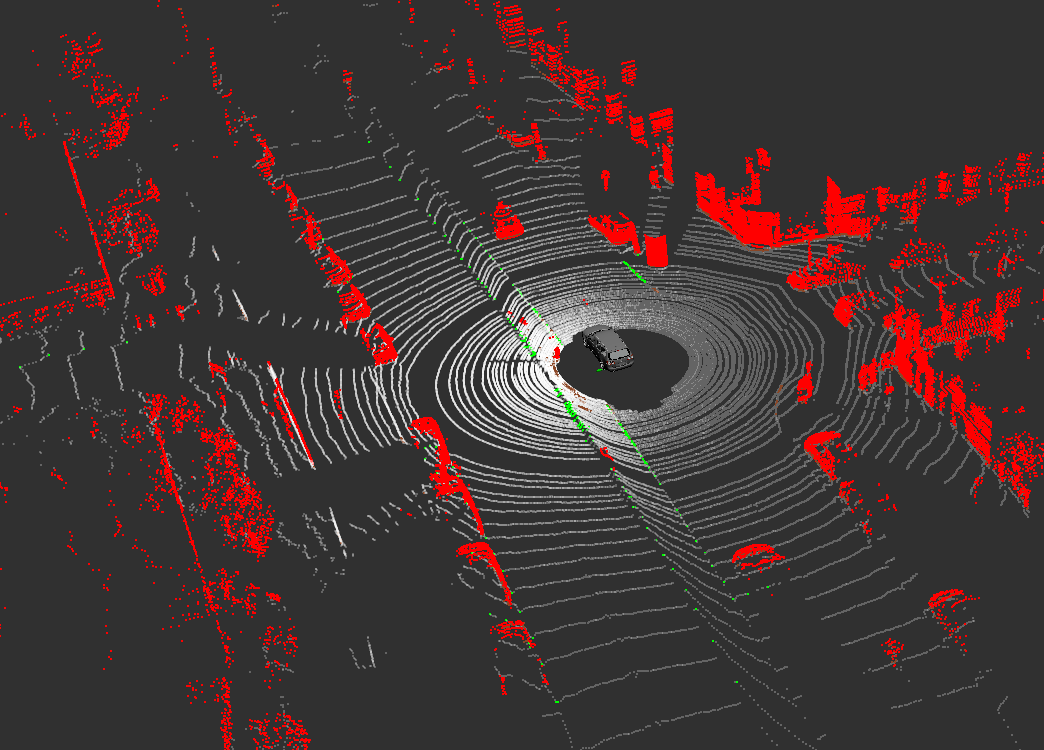}%
    } \hfill  
    \subfloat[Pointcloud clustering]{\label{subfig:Preprocessing_Example_Segmentation}%
        \includegraphics[width=0.32\textwidth]{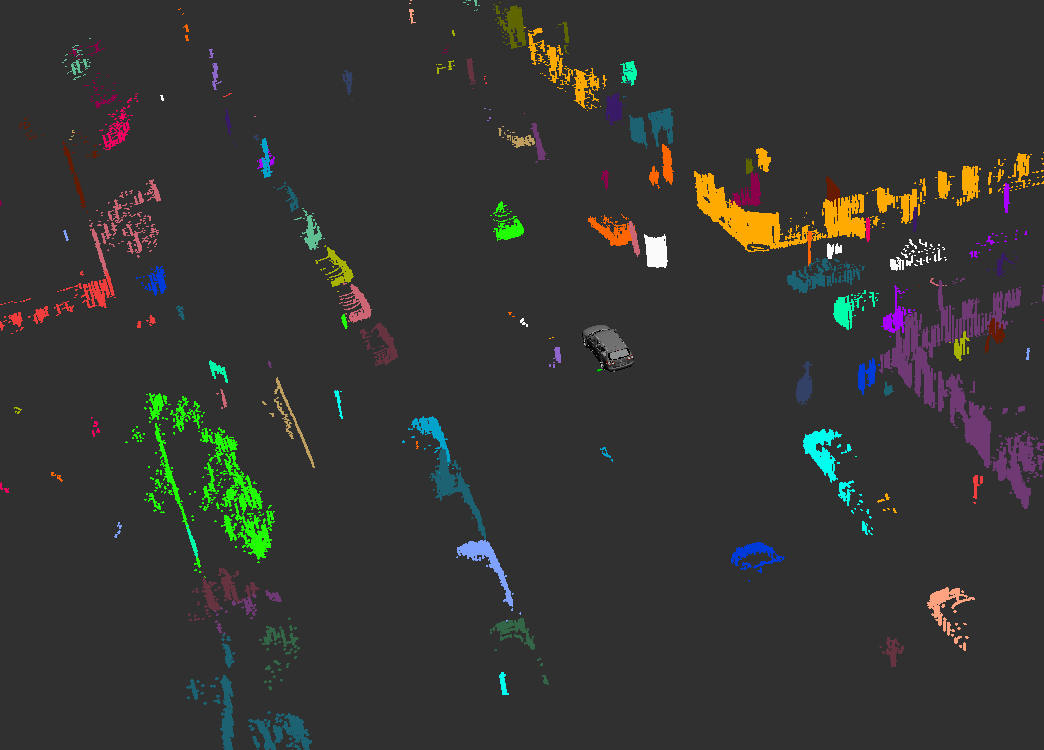}%
    }\hfill
    \subfloat[Motion class estimation]{\label{subfig:Preprocessing_Example_Dynclass}%
        \includegraphics[width=0.32\textwidth]{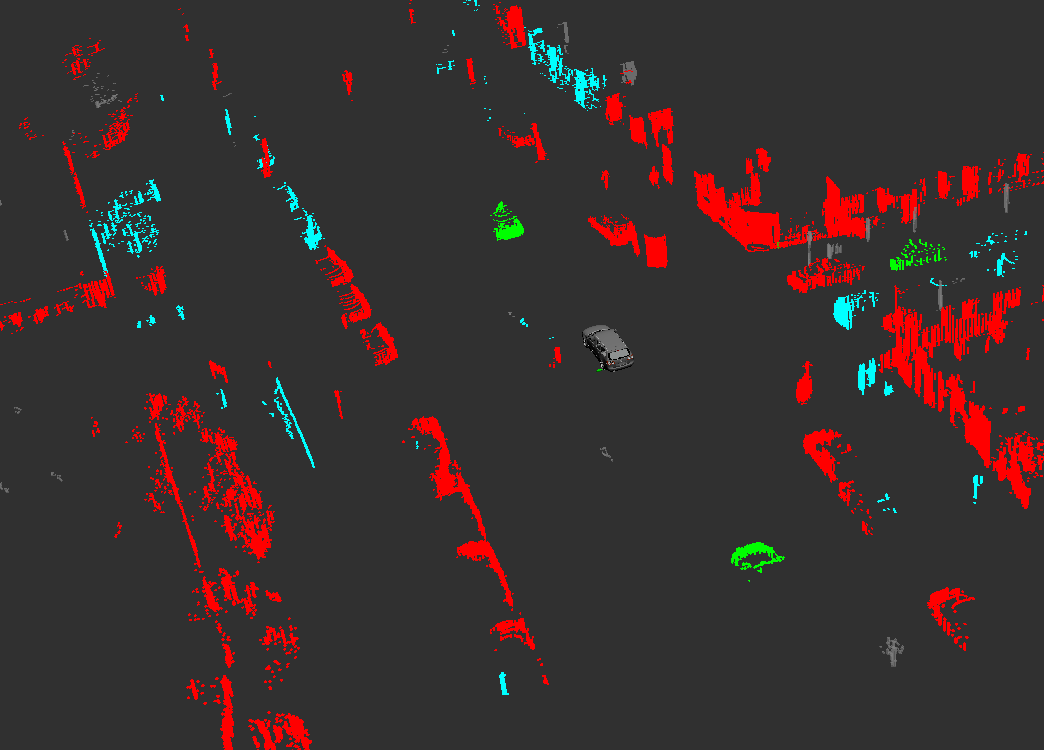}%
    }
    \caption{
        Results of the different preprocessing stages. 
        \protect\subref{subfig:Preprocessing_Example_Ground_Surface} illustrates the results of the ground surface estimation and curb detection (white: ground, red: elevated targets, green: curb structures). 
        The results of the cluster stage based on a Multi-Volume representation is shown in~\protect\subref{subfig:Preprocessing_Example_Segmentation}. 
        The same clusters are visualized by their motion class in \protect\subref{subfig:Preprocessing_Example_Dynclass} (green: movable, red: stationary, cyan: potentially movable).
    }
    \label{fig:Preprocessing_results}
    \vspace{-0.5em}
\end{figure*}

\subsection{Ground surface estimation}
\label{ssec:Preprocessing_Ground_Surface_Classification}
The first processing stage estimates the ground surface to separate between drivable areas and non-traversable obstacles.
Although assumed in many approaches, the ground surface is generally not a single plane, but has an arbitrary structure.
Additionally, protruding and overhanging structures need to be distinguishable from the ground surface in order to e.\,g. separate a car driving on the road even within a tunnel or beneath a tree.
While many approaches treat ground measurements as noise and focus on their removal, explicit representation of the ground surface yields beneficial information about drivable regions.   
Common approaches are either based on vertical displacement analysis in intermediate grid-based representations \citep{Thrun_Stanley_2006, Himmelsbach_perception_2009, Asvadi_Tracking_2015}. 
Others exploit the availability of adjacent scan lines from multi-beam LiDARs to classify the ground surface based on a slope threshold.
This can either be performed on a point-pair basis \citep{Petrovskaya_Vehicle_Detection_Tracking_2009, Montemerlo_DARPA_2008}, or by taking larger point clusters into account \citep{Himmelsbach_Segmentation_2010, Leonard_DARPA_2008}.
Additionally, probabilistic models are introduced \citep{Chen_Ground_Segmentation_2011,  Rummelhard_Ground_Segmentation_2017}.
While algorithms of the first group are usually not able to model overhanging objects properly, algorithms of the second group are vulnerable to sporadic erroneous measurements.
Yet, both groups yield short execution times.
Due to their underlying optimization procedure, algorithms of the third group tend to be computationally expensive.

To overcome the disadvantages of the first two groups while preserving their low computational demands, we developed a multi-step algorithm \citep[cf.][]{Rieken_Ground_Surface_2015}.
Measurements are pre-classified by a slope-based approach.
Those potential ground measurements are inserted in a grid-based structure, in which each cell models the height of the respective part of the environment. 
Median filtering is applied to this grid to filter occasional false detections.
In addition to above's method \citep{Rieken_Ground_Surface_2015}, we add an iterative region-fill algorithm to propagate information about the ground surface height to regions that do not contain pre-classified measurements.
The final measurement classification evaluates the vertical distance between the measurement itself and the height value of the respective grid cell.
\autoref{subfig:Preprocessing_Example_Ground_Surface} shows a typical result of this preprocessing stage.

\subsection{Curb detection}
Within the pointcloud, curb-like structures are detected to provide input for road boundary extraction algorithms.
Curb detection algorithms can be categorized similar to those for ground surface estimation and share the same algorithmic approaches, such as elevation mapping \citep{Oniga_Elevation_Mapping_2007} or (horizontal) slope detection \citep{Patz_DARPA_2008}.
\citet{Kellner_Curb_Detection_2014_ITSC} compare several approaches for segmenting single scan lines.
Due to its capability to deal with noisy measurements, their approach of fitting piece-wise line segments is adapted to our system.
It is extended by an angular threshold and a point density criterion.

\subsection{Vertical compression of the pointcloud}
In the following, we focus on the elevated elements.
We introduce a vertical compression to further reduce the total number of measurements for the subsequent processing stages.
The key idea is to use a grid structure to combine measurements with similar coordinates into single volumes, as presented by e.\,g. \citet{Pfaff_Extended_Elevation_Maps_2006}.
In contrast to elevation mapping techniques, which use a single height value per cell, a volumetric representation is able to consider overhanging structures.
\citet{Triebel_MLSP_2006} extended this approach to multiple volumes per cell.

To preserve the sensor's angular characteristics, we employ a polar grid structure with an angular discretization equal to the sensor's horizontal resolution~$\Delta \varphi_S$ and a radial discretization~$\Delta r$.
Within each cell, a vertical stack models occupancy at different height intervals.
The stack is implemented as a bit-field, leading to only minor computational overhead compared to a full 3D grid.
\begin{figure}
	\centering
	\tdplotsetmaincoords{75}{-20}
\colorlet{segment_color_1}{tuOrangeLight100}
\colorlet{segment_color_2}{tuBlueLight100}
\colorlet{segment_color_3}{tuGreenLight100}
\colorlet{segment_color_4}{tuVioletLight80}
\begin{tikzpicture}[tdplot_main_coords, scale=0.95]

	\tikzstyle{grid_outer_lines}=[black, solid];
	\tikzstyle{grid_help_lines} =[black!25, dashed];
	\tikzstyle{grid_arc_lines} =[grid_outer_lines, blue!50!black];

	\tikzstyle{cube_side}=[line cap=round, line width=.25pt];

	\tikzstyle{object_1_cube} =[fill=segment_color_1];
	\tikzstyle{object_2_cube} =[fill=segment_color_2];
	\tikzstyle{object_3_cube} =[fill=segment_color_3];
	\tikzstyle{object_4_cube} =[fill=segment_color_4];
	\tikzstyle{empty_cube}    =[fill=white, densely dotted];
	
	\tikzset{
		pics/cube/.style args ={#1:#2:#3:#4}{code= { 
			\draw[cube_side, #4] (0,0,#3*#2*0.5)-- ++(#1:#2) -- ++(0,0,#2-0.5*#2) -- ++(#1:-#2) -- (0,0,#3*#2*0.5);
			\draw[cube_side, #4] (0,0,#3*#2*0.5)-- ++(#1+90:#2) -- ++(0,0,#2-0.5*#2) -- ++(#1+90:-#2) -- (0,0,#3*#2*0.5);
			\draw[cube_side, #4] (0,0,#2*0.5+#3*#2*0.5) -- ++(#1:#2) -- ++(#1+90:#2) -- ++(#1:-#2);	
	}},
	}

	\pgfmathsetmacro{\radius}{7}
	\pgfmathsetmacro{\radiusminusone}{\radius - 1}
	
	\pgfmathsetmacro{\radiusstepsize}{1}
	\pgfmathsetmacro{\radiushelpergriddistance}{\radius - \radiusstepsize}
	\pgfmathsetmacro{\height}{2.5}

	\pgfmathsetmacro{\anglebegin}{0}
	\pgfmathsetmacro{\angleend}{40}
	\pgfmathsetmacro{\anglestepsize}{10}
	\pgfmathsetmacro{\anglesteps}{(\angleend - \anglebegin) /\anglestepsize - 1}
	
	\pgfmathsetmacro{\anglebracebegin}{\anglebegin + \anglestepsize}
	\pgfmathsetmacro{\anglebraceend}{\anglebracebegin + \anglestepsize}
	
	\pgfmathsetmacro{\cubesize}{0.33}
	\pgfmathsetmacro{\numcubes}{\height / \cubesize}

	\draw [grid_arc_lines] (0, 0, 0	    ) ++ (\anglebegin:\radius) arc (\anglebegin:\angleend:\radius);  
	
	\foreach \x [evaluate=\x as \anglevalue using \anglebegin + \x * \anglestepsize] in {1,...,\anglesteps}{
		\draw [grid_help_lines] (0, 0, 0) -- (\anglevalue:\radius);	
	}
	\foreach \x [evaluate=\x as \rangevalue using \x * \radiusstepsize] in {1,...,\radiushelpergriddistance}{
		\draw [grid_help_lines]  (0, 0, 0) ++ (\anglebegin:\rangevalue) arc (\anglebegin:\angleend:\rangevalue); 	
	}
	
	
	
	\draw [grid_outer_lines]      (0, 0, 0)       	-- (\anglebegin:\radius);
	\draw [grid_outer_lines]      (0, 0, 0)       	-- (\angleend:\radius);

	\draw [decorate, decoration={brace,amplitude=3pt, mirror}] (0:\radiusminusone) -- (0:\radius) node [midway, below,yshift=-0.2em] {$\Delta r = \SI{20}{\centi\meter}$}; 
	\draw [decorate, decoration={brace,amplitude=3pt, mirror}] (\anglebracebegin:\radius) -- (\anglebraceend:\radius) node [midway, above,xshift=2em] {$\Delta \varphi_S (\approx \ang{0.18})$};
	
	\pgfmathsetmacro{\stackrange}{3.25}
	\pgfmathsetmacro{\stackangle}{22.5}
	
	\draw [decorate, decoration={brace,amplitude=1.0pt}] 
			(0, 0.25, 5. * \cubesize*0.5)
		++  (\stackangle:\stackrange)
		++  (\stackangle+90:\cubesize) 
		--  ++(0,0, \cubesize*0.5) node [midway, left] {$\Delta z = \SI{12,5}{\centi\meter}$};
	
	\draw (\stackangle:\stackrange) pic {cube=\stackangle:\cubesize:0:object_1_cube};
	\draw (\stackangle:\stackrange) pic {cube=\stackangle:\cubesize:1:object_1_cube};
	\draw (\stackangle:\stackrange) pic {cube=\stackangle:\cubesize:2:object_1_cube};
	\draw (\stackangle:\stackrange) pic {cube=\stackangle:\cubesize:2:object_1_cube};
	\draw (\stackangle:\stackrange) pic {cube=\stackangle:\cubesize:3:empty_cube};
	\draw (\stackangle:\stackrange) pic {cube=\stackangle:\cubesize:4:empty_cube};
	\draw (\stackangle:\stackrange) pic {cube=\stackangle:\cubesize:5:empty_cube};
	\draw (\stackangle:\stackrange) pic {cube=\stackangle:\cubesize:6:empty_cube};
	\draw (\stackangle:\stackrange) pic {cube=\stackangle:\cubesize:7:empty_cube};	
	\draw (\stackangle:\stackrange) pic {cube=\stackangle:\cubesize:8:empty_cube};
	\draw (\stackangle:\stackrange) pic {cube=\stackangle:\cubesize:9:empty_cube};
	\draw (0,0, 0.85 *\height) ++ (\stackangle:\stackrange) node {$A$};

    \pgfmathsetmacro{\stackrange}{5.25}
    \pgfmathsetmacro{\stackangle}{32.5}
	\draw (\stackangle:\stackrange) pic {cube=\stackangle:\cubesize:0:empty_cube};
	\draw (\stackangle:\stackrange) pic {cube=\stackangle:\cubesize:1:empty_cube};
	\draw (\stackangle:\stackrange) pic {cube=\stackangle:\cubesize:2:object_2_cube};
	\draw (\stackangle:\stackrange) pic {cube=\stackangle:\cubesize:2:object_2_cube};
	\draw (\stackangle:\stackrange) pic {cube=\stackangle:\cubesize:3:object_2_cube};
	\draw (\stackangle:\stackrange) pic {cube=\stackangle:\cubesize:4:object_2_cube};
	\draw (\stackangle:\stackrange) pic {cube=\stackangle:\cubesize:5:object_2_cube};
	\draw (\stackangle:\stackrange) pic {cube=\stackangle:\cubesize:6:object_2_cube};
	\draw (\stackangle:\stackrange) pic {cube=\stackangle:\cubesize:7:empty_cube};	
	\draw (\stackangle:\stackrange) pic {cube=\stackangle:\cubesize:8:empty_cube};
	\draw (\stackangle:\stackrange) pic {cube=\stackangle:\cubesize:9:empty_cube};
	\draw (0,0, 0.85 *\height) ++ (\stackangle:\stackrange) node {$B$};

	\pgfmathsetmacro{\stackrange}{5.25}
    \pgfmathsetmacro{\stackangle}{2.5}
	\draw (\stackangle:\stackrange) pic {cube=\stackangle:\cubesize:0:empty_cube};
	\draw (\stackangle:\stackrange) pic {cube=\stackangle:\cubesize:1:object_3_cube};
	\draw (\stackangle:\stackrange) pic {cube=\stackangle:\cubesize:2:object_3_cube};
	\draw (\stackangle:\stackrange) pic {cube=\stackangle:\cubesize:2:object_3_cube};
	\draw (\stackangle:\stackrange) pic {cube=\stackangle:\cubesize:3:object_3_cube};
	\draw (\stackangle:\stackrange) pic {cube=\stackangle:\cubesize:4:empty_cube};
	\draw (\stackangle:\stackrange) pic {cube=\stackangle:\cubesize:5:empty_cube};
	\draw (\stackangle:\stackrange) pic {cube=\stackangle:\cubesize:6:empty_cube};
	\draw (\stackangle:\stackrange) pic {cube=\stackangle:\cubesize:7:object_4_cube};	
	\draw (\stackangle:\stackrange) pic {cube=\stackangle:\cubesize:8:object_4_cube};
	\draw (\stackangle:\stackrange) pic {cube=\stackangle:\cubesize:9:object_4_cube};
	\draw (0,0, 0.85 *\height) ++ (\stackangle:\stackrange+0.1) node  {$C$};

	\draw[-stealth', semithick] (0, 0, 0) -- ++(1, 0, 0) node [below] {$r$};
	\draw[-stealth', semithick] (0, 0, 0) -- ++(0, 0, 1) node [left] {$z$};
	\draw [-stealth', semithick](0, 0, 0) ++(0:1.0) arc (0:\angleend:1.0) node [near end, above left] {$\varphi$}; 

\end{tikzpicture}
    \vspace*{-1em}
	\caption{
		Pointcloud compression stage. 
		Each polar grid cell contains a vertical stack that is populated by the current pointcloud data.
		In each stack, neighboring occupied cells are combined to one compressed measurement with a minimum height and a vertical extent. 
		The example shows the capability to represent ground-connected elements ($\text{A}$), levitating elements ($\text{B}$), and multiple structures per stack ($\text{C}$).
	}
    \vspace{-1em}
	\label{fig:Preprocessing_Histogram_Bins}
\end{figure}
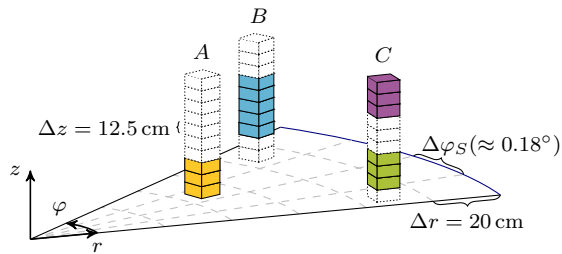

All non-ground measurements are projected into the vertical stacks, as illustrated in \autoref{fig:Preprocessing_Histogram_Bins}.
Neighboring occupied cells within such a stack are condensed to a compressed measurement~$\vec{p}^\prime$.
Using the minimal radial distance~$r_\text{min}$ of the assigned points, the compressed measurement is represented by $\vec{p}^\prime = \left(r_\text{min}, \varphi, z_\text{min}, h \right)^T$.
In case of multiple groups per stack, multiple volumes are created.
To cope with isolated false detections, we extend this process by density-based filtering techniques.
At least two measurements have to be assigned to one stack.
Alternatively, a minimum number of measurements has to be available within the surrounding of the current stack. 
Calculation of this neighborhood density is accomplished efficiently by using summed area tables \citep{Crow_Summed_Area_Tables_1984}.

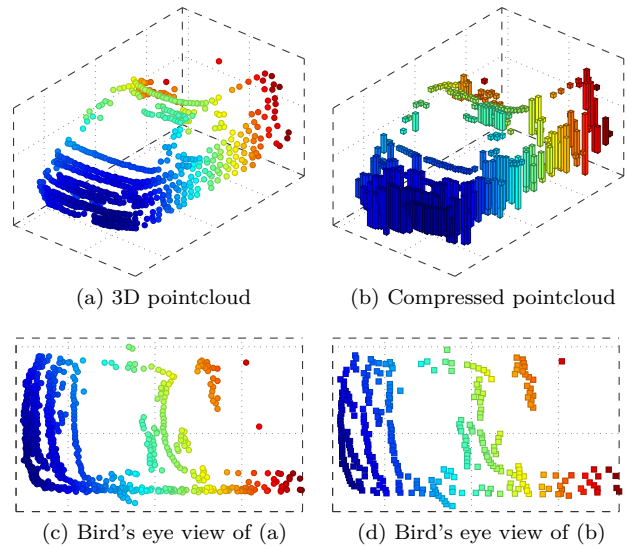
\begin{figure}
    \centering
    \tikzexternalenable
    \subfloat[3D pointcloud]{ \label{subfig:Preprocessing_Multi_Volume_Representation_3D_Points}
    \begin{tikzpicture}[scale=1]

\begin{axis}[
	width=6.5cm,
	anchor=origin,
	view/h=40,
	xmin=3.1, xmax=5.1,
	ymin=11.4, ymax=14.7,
	zmin=0, zmax=1.7,
	axis equal image,
	grid=both,
	grid style={line width=.2pt, draw=black!80, dotted},
	axis line style={black!80, dashed},
	xtick distance=1,
	ytick distance=1,
	ztick distance=1,
	ticks=none,
	x dir=reverse,	
	y tick label style={
        /pgf/number format/.cd,
            fixed,
            fixed zerofill,
            precision=1,
        /tikz/.cd
    },
    x tick label style={
        /pgf/number format/.cd,
            fixed,
            fixed zerofill,
            precision=1,
        /tikz/.cd
    },
    z tick label style={
        /pgf/number format/.cd,
            fixed,
            fixed zerofill,
            precision=1,
        /tikz/.cd
    },
    point meta min=145,
    point meta max=228
	]
 	\addplot3[
 		scatter,
 		only marks, 
 		mark options=solid, 
 		mark size=1pt,
 		] table [
 			point meta=explicit, 
 			x expr = \thisrow{y}, 
 			y expr = \thisrow{x}, 
 			z expr = \thisrow{z},
 			col sep=comma, 
 			meta expr={\thisrow{x}^2 + \thisrow{y}^2}
 		] {\currentimagedirectory/tikz_data_raw.csv};	  
 \end{axis}

\end{tikzpicture}
    }
    \subfloat[Compressed pointcloud]{  \label{subfig:Preprocessing_Multi_Volume_Representation_3D_Stixels}
        \begin{tikzpicture}[scale=1]

\pgfplotstableset{col sep=comma}
\pgfplotstableread{\currentimagedirectory/tikz_data_stixels.csv}\pointcloud
\pgfplotstablegetrowsof{\pointcloud}
\pgfmathsetmacro{\NumPoints}{\pgfplotsretval-1}

\xdef\rangemin{145.0}
\xdef\rangemax{228.0}

\xdef\rectwidth{0.05}
\xdef\rectlength{0.05}

    \tikzset{
        rangecolormap/.style={
            color of colormap={(#1 - \rangemin) / (\rangemax - \rangemin) * 1000.0},
           draw=.!50!black,
            fill=.,
            line width=0.05pt,
            line cap=round,
        },
	}
\begin{axis}[
	width=6.5cm,
	anchor=origin,
	view/h=40,
	xmin=3.1, xmax=5.1,
	ymin=11.4, ymax=14.7,
	zmin=0, zmax=1.7,
	axis equal image,
	grid=both,
	grid style={line width=.2pt, draw=black!80, dotted},
	axis line style={black!80,, dashed},
	xtick distance=1,
	ytick distance=1,
	ztick distance=1,
	ticks=none,
	x dir=reverse,	
	y tick label style={
        /pgf/number format/.cd,
            fixed,
            fixed zerofill,
            precision=1,
        /tikz/.cd
    },
    x tick label style={
        /pgf/number format/.cd,
            fixed,
            fixed zerofill,
            precision=1,
        /tikz/.cd
    },
    z tick label style={
        /pgf/number format/.cd,
            fixed,
            fixed zerofill,
            precision=1,
        /tikz/.cd
    },
	z buffer=sort
	]
		\foreach \row in {0,...,\NumPoints}{
			\pgfplotstablegetelem{\row}{x}\of{\pointcloud}
			\xdef\y{\pgfplotsretval}
			\pgfplotstablegetelem{\row}{y}\of{\pointcloud}
			\xdef\x{\pgfplotsretval}
			\pgfplotstablegetelem{\row}{z}\of{\pointcloud}
			\xdef\zmin{\pgfplotsretval}
			\pgfplotstablegetelem{\row}{height}\of{\pointcloud}
			\xdef\height{\pgfplotsretval}
			\xdef\range{\x * \x + \y * \y}
			\edef\pointdata{\noexpand %
				\fill[rangecolormap=\range] (\x - \rectwidth / 2.0, \y - \rectlength / 2.0, \zmin+\height)  
						-- ++(axis direction cs: \rectwidth, 0, 0)
						-- ++(axis direction cs: 0, \rectlength, 0)
						-- ++(axis direction cs: -\rectwidth, 0, 0) -- cycle; %
				}
			\pointdata
			\edef\pointdata{\noexpand %
				\fill[rangecolormap=\range] (\x - \rectwidth / 2.0, \y - \rectlength / 2.0, \zmin)  
						-- ++(axis direction cs: \rectwidth, 0, 0)
						-- ++(axis direction cs: 0, 0, \height)
						-- ++(axis direction cs: -\rectwidth, 0, 0) -- cycle; %
				}
			\pointdata
			\edef\pointdata{\noexpand %
				\fill[rangecolormap=\range] (\x - \rectwidth / 2.0, \y - \rectlength / 2.0, \zmin)  
						-- ++(axis direction cs: 0, \rectlength, 0)
						-- ++(axis direction cs: 0, 0, \height)
						-- ++(axis direction cs: 0, -\rectlength, 0) -- cycle; %
				}
			\pointdata
	};

 	\end{axis}

\end{tikzpicture}
    }

    \subfloat[Bird's eye view of \protect\subref{subfig:Preprocessing_Multi_Volume_Representation_3D_Points}]{
    \begin{tikzpicture}[scale=1]

\begin{axis}[
	width=6.25cm,
	anchor=origin,
	view/h=90,
	view/v=90,
	xmin=3.1, xmax=5.1,
	ymin=11.4, ymax=14.7,
	zmin=0, zmax=2,
	axis equal image,
	grid=both,
	grid style={line width=.2pt, draw=black!80, dotted},
	axis line style={black!80,, dashed},
	xtick distance=1,
	ytick distance=1,
	ztick distance=1,
	ticks=none,
	x dir=reverse,	
	y tick label style={
        /pgf/number format/.cd,
            fixed,
            fixed zerofill,
            precision=1,
        /tikz/.cd
    },
    x tick label style={
        /pgf/number format/.cd,
            fixed,
            fixed zerofill,
            precision=1,
        /tikz/.cd
    },
    z tick label style={
        /pgf/number format/.cd,
            fixed,
            fixed zerofill,
            precision=1,
        /tikz/.cd
    },
    point meta min=145,
    point meta max=228
	]
 	\addplot3[scatter, only marks, mark options=solid,  mark size=1pt,
 	] table [point meta=explicit, x expr= \thisrow{y}, y expr = \thisrow{x}, z expr =\thisrow{z}, col sep=comma, meta expr={\thisrow{x}^2 + \thisrow{y}^2}] {\currentimagedirectory/tikz_data_raw.csv};	  
 \end{axis}

\end{tikzpicture}
    }
    \subfloat[Bird's eye view of \protect\subref{subfig:Preprocessing_Multi_Volume_Representation_3D_Stixels}]{
        \begin{tikzpicture}[scale=1]

\begin{axis}[
	width=6.25cm,
	anchor=origin,
	view/h=90,
	view/v=90,
	xmin=3.1, xmax=5.1,
	ymin=11.4, ymax=14.7,
	zmin=0, zmax=2,
	axis equal image,
	grid=both,
	grid style={line width=.2pt, draw=black!80, dotted},
	axis line style={black!80,, dashed},
	xtick distance=1,
	ytick distance=1,
	ztick distance=1,
	ticks=none,
	x dir=reverse,	
	y tick label style={
        /pgf/number format/.cd,
            fixed,
            fixed zerofill,
            precision=1,
        /tikz/.cd
    },
    x tick label style={
        /pgf/number format/.cd,
            fixed,
            fixed zerofill,
            precision=1,
        /tikz/.cd
    },
    z tick label style={
        /pgf/number format/.cd,
            fixed,
            fixed zerofill,
            precision=1,
        /tikz/.cd
    },
    point meta min=145,
    point meta max=228
	]
 	\addplot3[	scatter,
				scatter/use mapped color={draw=mapped color!75!black, fill=mapped color}, 	
 				only marks,
 				mark=square*,
 				mark size=1pt,
 				mark options={solid, line width=0.03pt},
 				] 
 				table [
 					point meta=explicit,
 					x expr = \thisrow{y},
 					y expr = \thisrow{x},
 					z expr = \thisrow{z},
 					col sep=comma, 
 					meta expr={\thisrow{x}^2 + \thisrow{y}^2}
 				] 
 			 {\currentimagedirectory/tikz_data_stixels.csv};	  
 \end{axis}

\end{tikzpicture}
    }
    \tikzexternaldisable
    \caption{Example of pointcloud compression results, illustrated by the pointcloud \protect\subref{subfig:Preprocessing_Multi_Volume_Representation_3D_Points} and its corresponding compression result \protect\subref{subfig:Preprocessing_Multi_Volume_Representation_3D_Stixels}. 
        Points are colored by their radial distance.
}
    \label{fig:Preprocessing_Multi_Volume_Representation}
    \vspace{-1em}
\end{figure}

As a result, the number of measurements is reduced to \num{25}-\SI{50}{\percent}, depending on the amount of vertical structures present in the current pointcloud.
Compression and radial discretization lead to removal of details (see \autoref{fig:Preprocessing_Multi_Volume_Representation}).
However, the outer contours of structures remain intact.

\subsection{Pointcloud clustering} 
Clustering of the compressed measurements uses similarity criteria to define whether some given points belong to the same physical object\footnote{In the context of this work, the separation of \emph{physical objects} is the main focus. In other domains, different criteria might be more suitable.}.
During the last decades, many approaches dealing with clustering of pointcloud data were presented.
They can be categorized by the type of features they use for defining this similarity.
Similarity criteria are usually defined by some distance metric applied to the feature space of the measurements,
hence both terms are often considered in conjunction.

\subsubsection{Similarity criteria and distance metrics}
The most predominant similarity criterion is the spatial neighborhood between measurements.
Next to this, one popular approach is the additional incorporation of a density criterion for dealing with noisy measurements, e.\,g. the \emph{Density Based Spacial Clustering of Applications with Noise} (\emph{DBSCAN}) approach by \citet{Ester_DBSCAN_1996}.
In addition, clustering may be performed on derived features, like normal vectors \citep{Klasing_Segmentation_2009}, local convexity \citep{Moosmann_Segmentation_2009} or based on geometric features like line structures \cite{Zhao_Tracking_1998, Kellner_Curb_Detection_2014_ITSC}.
Another option is to incorporate additional data sources into the clustering process. 
Those may be information about already tracked movable objects \citep{Himmelsbach_Tracking_Detection_2012}, results from motion classification (see \autoref{ssec:Motion_Classification}), but also class information from camera images \citep{Held_Segmentation_2016}.

\subsubsection{Techniques for fast neighbor search}
Usually, clustering algorithms are based on finding neighboring points in the given measurements. 
In case of unordered measurements, this requires the computationally expensive calculation of the distance metric to all other measurements.
Hence, many techniques were applied to speed up this part.
\citet{Klasing_Segmentation_2008} use a \emph{kd-tree} as indexing structure.
Grid structures as intermediate representation are applicable as well, as they provide an implicit neigh\-bor\-hood definition by the cells' positions \citep{Steinhauser_Motion_Segmentation_2008, Himmelsbach_Segmentation_2010, Azim_Moving_Tracking_2012}.
\\
Apart from artificial indexing, scanning sensors' pointclouds contain an intrinsic order, which can be exploited for this purpose as well.
\citet{Sparbert_Segmentation_2001} present a fast clustering approach, which limits the number of required comparisons to only one neighboring channel. 
\citet{Klasing_Segmentation_2009} show an approach which calculates cluster features within a continuous measurement stream.
Next to these approaches, which are applied to single scan lines only, some work focuses on depth-image like  representations of multi-beam sensors \citep{Bogoslavskyi_Segmentierung_2016, Burger_Segmentierung_2018, Moosmann_Segmentation_2009, Zermas_Segmentation_2017}.
\begin{figure}
    \centering
        \subfloat[Height similarity criterion $\mathcal{E}_H$ (side view)]{
        \begin{minipage}[b]{0.48\textwidth}
            \centering
            \begin{tikzpicture}


\tikzstyle{heightextension}=[black!50, dashed, semithick];
\tikzstyle{heightline}=[black!85, dotted];

\pgfmathsetmacro{\heightthreshold}{0.4}
\pgfmathsetmacro{\upperstixelzmin}{1.5}
\pgfmathsetmacro{\upperstixelheight}{1.0}
\pgfmathsetmacro{\upperstixelradius}{4}
\pgfmathsetmacro{\lowerstixelradius}{1.5}
\pgfmathsetmacro{\stixelwidth}{0.33}


\pgfdeclarelayer{foreground}
\pgfsetlayers{main,foreground}

\begin{axis}
	[
		xmin = 0, xmax= 6,
		ymin = 0, ymax= 3,
		width = 8cm,
		axis equal image,	
		grid=none,
		grid style={line width=.25pt, draw=black!80, dotted},
		axis lines=middle,
		axis line style={black!80, -stealth', dashed},		
		xlabel={$r$},
		ylabel={$z$},
		xtick distance=1,
		ytick distance=1,
		ticks=none,
		x label style={anchor=north east},
		y label style={anchor=south west},
	]

\draw[heightline] (0, \upperstixelzmin + \upperstixelheight + \heightthreshold) -- ++(4, 0);
\draw[heightline] (0, \upperstixelzmin - \heightthreshold) -- ++(4, 0);

\draw[heightextension, segment_color_1] (\lowerstixelradius, 0.5) rectangle ++(\stixelwidth, -\heightthreshold);
\draw[heightextension, segment_color_1] (\lowerstixelradius, 1.5) rectangle ++(\stixelwidth, \heightthreshold);

\draw[heightextension, segment_color_2] (\upperstixelradius, \upperstixelzmin - \heightthreshold) rectangle ++(\stixelwidth, \heightthreshold);
\draw [decorate, decoration={brace,amplitude=3.5pt, mirror}] (\upperstixelradius + \stixelwidth, \upperstixelzmin - \heightthreshold) -- ++(0, \heightthreshold) node[midway, right, black] {$\Delta h_\text{max}$};

\draw[heightextension, segment_color_2] (\upperstixelradius, \upperstixelzmin + \upperstixelheight) rectangle ++(\stixelwidth, \heightthreshold);
\draw [decorate, decoration={brace,amplitude=3.5pt, mirror}] (\upperstixelradius + \stixelwidth, \upperstixelzmin + \upperstixelheight) -- ++(0, \heightthreshold) node[midway, right, black] {$\Delta h_\text{max}$};

\draw[dotted, tubsGray] (\lowerstixelradius + 0.5 * \stixelwidth, 0.5 ) -- ++(0,-0.5) node (R1End) {};
\draw [fill=segment_color_1] (\lowerstixelradius, 0.5) rectangle ++(\stixelwidth, 1);
\draw [decorate, decoration={brace,amplitude=3.5pt}] (\lowerstixelradius, 0.5) -- ++(0, 1) node[midway, left] {$h_n$};
\draw (\lowerstixelradius + \stixelwidth, 0.5 ) node [right] {$z_{n, \text{min}}$};

\draw[dotted, tubsGray] (\upperstixelradius + 0.5 * \stixelwidth, \upperstixelzmin ) -- ++(0,-\upperstixelzmin) node (R2End) {};

\draw [fill=segment_color_2] (\upperstixelradius, \upperstixelzmin) rectangle ++(\stixelwidth, \upperstixelheight);
\draw [decorate, decoration={brace,amplitude=3.5pt, mirror}] (\upperstixelradius + \stixelwidth, \upperstixelzmin) -- ++(0, \upperstixelheight) node[midway, right] {$h_m$};


\begin{pgfonlayer}{foreground}
	\node[below] at (R1End) {$r_n$};
	\node[below] at (R2End) {$r_m$};
\end{pgfonlayer}
\end{axis}
\end{tikzpicture}
            \label{subfig:Segmentation_Height_Similarity}
        \end{minipage}
    }

    \subfloat[Radial distance criterion $\mathcal{E}_R$ (bird's eye view)]{
        \begin{minipage}[b]{0.48\textwidth}
            \centering
            \label{subfig:Segmentation_Radial_distance_criteria}

\tdplotsetmaincoords{0}{-0}
\begin{tikzpicture}[tdplot_main_coords]

	\tikzstyle{grid_outer_lines}=[black, dashed];
	\tikzstyle{grid_help_lines} =[black!25, dotted];
	\tikzstyle{grid_arc_lines} =[grid_outer_lines, blue!50!black];
	\tikzstyle{distance_lines} =[black!75, dashed];
	\tikzstyle{limit_lines} =[solid, line width=1.5pt, black!50];

	\pgfmathsetmacro{\maxradius}{7}
	\pgfmathsetmacro{\anglebegin}{0}
	\pgfmathsetmacro{\angleend}{30}
	\pgfmathsetmacro{\anglestepsize}{7.5}
	\pgfmathsetmacro{\anglesteps}{(\angleend - \anglebegin) /\anglestepsize - 2}

	\pgfmathsetmacro{\radiusstepsize}{1}
	\pgfmathsetmacro{\radiushelpergriddistance}{\maxradius - \radiusstepsize}

	\draw [grid_arc_lines] (0, 0) ++ (\anglebegin:\maxradius) arc (\anglebegin:\angleend:\maxradius);  
	
	\foreach \x [evaluate=\x as \anglevalue using \anglebegin + \x * \anglestepsize] in {1,...,\anglesteps}{
		\draw [grid_help_lines] (0, 0) -- (\anglevalue:\maxradius);	
	}
	\foreach \x [evaluate=\x as \rangevalue using \x * \radiusstepsize] in {1,...,\radiushelpergriddistance}{
		\draw [grid_help_lines]  (0, 0) ++ (\anglebegin:\rangevalue) arc (\anglebegin:\angleend:\rangevalue); 	
	}
	
	\draw [grid_outer_lines, -stealth']      (0, 0)       	-- (\anglebegin:\maxradius) node [below] {$r$};
	\draw [grid_outer_lines, -stealth']      (0, 0) ++ (\anglebegin:\maxradius) arc (\anglebegin:\anglestepsize:\maxradius) node [midway, right] {$\varphi$};
	\draw [grid_outer_lines]      (0, 0)       	-- (\angleend:\maxradius);
	
	\coordinate (secondpoint) at (19:6);
	\coordinate (limitpoint) at (19:\maxradius);
	\coordinate (limitpointprojected) at (\anglebegin:4);
	\coordinate (origin) at (0,0);
	\coordinate (firstpoint) at ($(limitpoint)!0.6!(limitpointprojected)$);

	\draw[distance_lines, segment_color_1] (origin) -- (firstpoint)  node [near end, above, black] {$r_n$};
	\draw[distance_lines, segment_color_2] (origin) -- (secondpoint) node [near end, above, black] {$r_m$};
	\draw[distance_lines]                  (secondpoint) -- (limitpoint);
	
	\draw[limit_lines] (limitpoint) -- (firstpoint);
	\draw[decorate, decoration={brace,amplitude=8pt}] (limitpoint) -- (firstpoint) node [midway, below, anchor=north west, yshift=-0.5em, xshift=-0.5em] {$\,\, s_1 \cdot r_n$};
	\draw[limit_lines] (limitpointprojected) -- (firstpoint);
	
	\pic [pic text = {$\lambda_\text{max}$},draw, -stealth', black, angle radius = 1.05cm,
  angle eccentricity = 1.5] {angle = origin--firstpoint--limitpointprojected};
	\node [below] at (limitpointprojected) {};
	\draw[stealth'-stealth', distance_lines, black] (firstpoint) -- (secondpoint) node[midway, left] {$d$};	
	
	\pic [pic text={$\Delta\varphi_S$}, draw, -stealth', black, angle radius = 2.3cm,
  angle eccentricity = 1.2] {angle = firstpoint--origin--secondpoint};

	\draw[fill = segment_color_1] (firstpoint) circle (2.5pt) node [right, anchor=north west, yshift=0.2em] {$\vec{p}_n$};
	\draw[fill = segment_color_2] (secondpoint) circle (2.5pt) node [above] {$\vec{p}_m$};
	
	\draw[black, fill = black!50, solid] (limitpoint) circle (2.5pt) node [above right] {$\vec{p}_m^*$};
\end{tikzpicture}
        \end{minipage}
    }
    \caption{
        Similarity criteria for pointcloud clustering.  
        A height similarity criterion \protect\subref{subfig:Segmentation_Height_Similarity} is applied to account for vertical overlaps of neighboring points. 
        Two points are assigned to the same cluster if their vertical extent overlaps or is within a threshold~$\Delta h_\text{max}$.
        \protect\subref{subfig:Segmentation_Radial_distance_criteria} shows the radial distance criterion for assigning point~$\vec{p}_n$ to the same cluster as $\vec{p}_m$.
        Therefore, the distance~$d$ is compared to the distance $s_1 \cdot r_n$, which is derived from an angular threshold $\lambda_\text{max}$. 
        The point~$\vec{p}_m^*$ marks the upper distance limit for a positive assignment. 
    }    
    \label{fig:Preprocessing_Segmentation_Criteria}
    \vspace{-1.0em}
\end{figure}
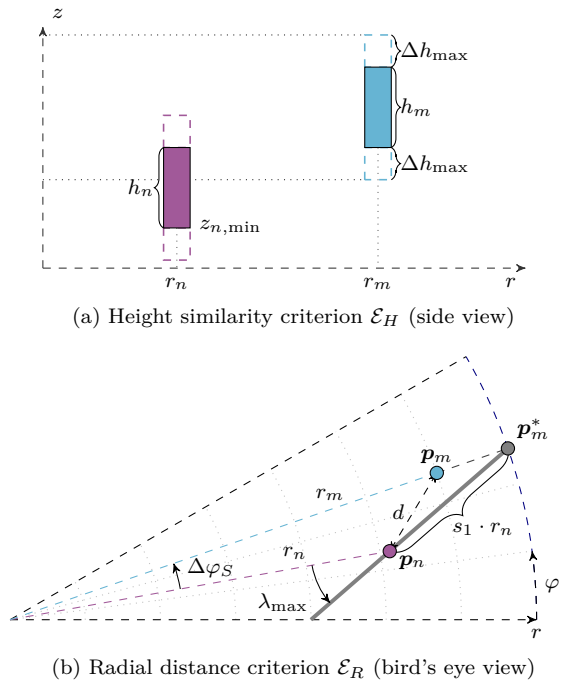

\subsubsection{Our approach}
Instead of using artificial indexing, the intrinsic ordering of the pointcloud is exploited in our project.
We extend the algorithm of \citet{Sparbert_Segmentation_2001} to consider height information of the neighboring cells and furthermore to handle multiple targets per angular step.
The latter is an important extension to solely depth-image based approaches like \citep{Bogoslavskyi_Segmentierung_2016, Moosmann_Segmentation_2009}.
This feature is required to deal with multiple volume elements per angle as provided by the pointcloud compression stage.

Starting with an arbitrary channel, a measurement $\vec{p}_n = \left(r_n, \varphi_n, z_\text{min}, h_n\right)^T$ is compared to the measurements within a limited number of adjacent preceding channels.
For each measurement~$\vec{p}_m$ in this neighborhood, two similarity criteria~$\mathcal{E}_R$ and~$\mathcal{E}_H$ need to be fulfilled (\autoref{fig:Preprocessing_Segmentation_Criteria}).
For $\mathcal{E}_R$, the radial distance of two points is given by:
\begin{align}
\function{d} \left(r_n, r_m\right) &= \sqrt{r_n^2 + r_m^2 - 2 r_n r_m \cos\left(\Delta\varphi_S\right) } \\ &\approx \left| r_n - r_m\right| \nonumber
\end{align}
This holds for small angular channel steps~$\Delta\varphi_S$ of the sensor system.
The distance $\function{d}$ is compared to a range-dependent threshold $\function{s}$, which is defined by a constant noise threshold~$s_0$ and a linear factor $s_1$:
\begin{align} \label{eq:Segmentation_Radial_Distance}
    \function{s} \left( r_n, r_m \right) &= s_0 + \min\left(r_n, r_m\right) \cdot \underbrace{ \frac{\sin\left(\Delta\varphi_S\right)}{\sin\left(\lambda_\text{max} - \Delta\varphi_S\right)}}_{= s_{1}}  
\end{align}
Here, $s_1$ is based on the \emph{Adaptive Breakpoint Detector} approach of \citet{Borges_Adaptive_Breakpoint_detector_2004}.
It defines a threshold angle~$\lambda_\text{max}$ relative to the current viewing direction.
This threshold can be interpreted as the maximum angle of a line structure whose detections in neighboring channels are required to be grouped into one cluster.

For $\mathcal{E}_H$, the height components of~$\vec{p}_n$ and~$\vec{p}_m$ are checked for overlap.
A tolerance~$\Delta h_\text{max}$ is applied to deal with occasionally missing measurements (see \autoref{subfig:Segmentation_Height_Similarity}).

For positive similarity results, labels are assigned via a \emph{Connected Components Labeling} algorithm. 
It provides an intrinsic solution for merging previously separated clusters. 
The results of the clustering algorithm are illustrated in \autoref{subfig:Preprocessing_Example_Segmentation}.

\subsection{Consistency-based motion classification}
\label{ssec:Motion_Classification}
The motion classification step separates movable from stationary elements.
We use a consistency grid approach, like presented by \citet{Matthaei_Consistency_FUSION_2011, Matthaei_Grid_Road_Detection_2013}.
A finite state machine per grid cell infers motion states, based on each cell's free and hit update history.
The cells' states are evaluated and forwarded to the measurements of the current pointcloud.
Next to the classes \emph{stationary} and \emph{movable}, the intermediate class \emph{potentially movable} is assigned to measurements with a yet ambiguous detection state. 
Finally, a single motion class per cluster is determined by a majority vote (see \autoref{subfig:Preprocessing_Example_Dynclass}).

\section{Stationary environment modeling}
\label{sec:Stationary_Environment_Modeling}

\providecommand{\currentimagedirectory}{}
\renewcommand{\currentimagedirectory}{./figures/5_Grids}

Up to this stage, we enhanced the pointcloud with information regarding the ground surface, the measurements' motion states and their assignment to clusters.
Yet, the pointcloud represents only the most recent scan without any temporal filtering or tracking and is thus prone to occasional false detections or inconclusive results.
To infer proper results for the context modeling stage, temporal filtering and tracking are accomplished by a dual representation, in which the stationary environment is modeled separately from the moving parts.
 
The stationary part is modeled via a multi-layer grid approach.
Early versions of this concept were introduced by \citet{Matthaei_Grids_2014}, which use multi-layer grids for combining multiple features for lane and road detection. 
In \citep{Rieken_Environment_Modelling_2015}, we extended this work towards a larger set of features, including information about explicitly detected traversable regions and curb structures.
The key idea of the approach is to model various features of the 3D stationary environment by a combination of 2D feature grids and create a combined semantic grid using a fusion scheme.
In the following section, the main aspects of the multi-layer grid approach are summarized and the latest results are presented.

\subsection{Multi-layer feature representation and grid fusion} 
In general, a single grid layer is not suitable for storing information from multiple sources, e.\,g. different sensors, due to different update rates, field of views, information qualities or information types.
Thus, we incorporate multiple layers for the different feature types.

The multi-layer grid stack currently consists of five different layers.
The elevated parts of the stationary environment are represented by the well-known Bayesian occupancy grids \citep{Elfes_Grids_1989}. 
The occupancy grid model is extended by a height value, similar to an elevation map.
Each cell stores the height information relative to the estimated ground height at the cell's position.
With two instances of this grid model, we are able to model the upper and lower bounds in vertical direction, which allows the representation of over- and underrunable regions. 
The ray-based inverse sensor model utilizes the pointcloud's motion classes.  
While all first-target measurements are taken into account for freespace update, only those with a class different from \emph{movable} are allowed to generate hit updates.
Hereby, we avoid that movable elements lead to artifacts in the grid, which would result from the fact that moving elements violate the grid model's assumption of a stationary environment.  
Yet, we can infer freespace up to the occurrence of movable elements.

Curb structures are represented by a regular Bayesian occupancy grid, yet with a different parametrization of range limits and update weights to properly consider the different feature detection characteristics.

Intensity information of the ground points is incorporated into a reflectance grid, as presented by \citet{Levinson_Dissertation_2011, Matthaei_Grids_2014}.
This grid represents textural properties of the road and ground surface. 

Finally, a layer stores information about regions with visible ground surface.
Hereby, traversable regions are represented explicitly and thus enable a conservative estimation of the currently available freespace. 
This is a conceptual difference from those approaches that derive traversable regions from the absence of occupancy, as usually done within Bayesian occupancy modeling.
Information about traversable regions is derived from the presence of ground information and represented by the last time instant at which this information was available.
Similar to the work of \citet{Glaser_Perception_2014}, ground information is inferred from the presence of ground measurements, but is also interpolated between neighboring pointcloud layers.
This is accomplished by extending the freespace modeling approach from \citet{Yu_Evidential_Sensor_Model_2014} with the results of our ground surface estimation.

Having specific features stored in the different layers, their cell values are converted into a tristate representation (i.\,e., indicating that a feature is \emph{existent}, \emph{non-existent}, or \emph{indecisive}).
The tristate values are fused cell-wise into a single layer by a rule-based approach.
This \emph{fusion layer} provides a combined data representation which may be interpreted as a \emph{semantic} view of the stationary environment.
\autoref{subfig:Results_Fusion_Grid}  shows the results in a real-world scene.

\subsection{Feature extraction}
\begin{figure}
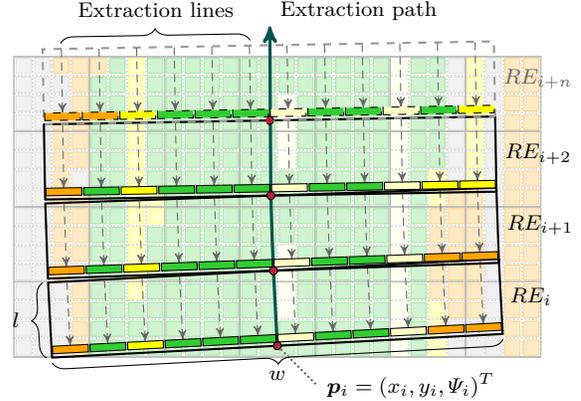

    \centering
    \tikzexternalenable
    \begin{tikzpicture}

\input{\currentimagedirectory/grid_celltypes_colormaps.tikz}
\input{\currentimagedirectory/grid_pic_definitions.tikz}

\pgfmathsetmacro{\elementwidth}{6}
\pgfmathsetmacro{\elementlength}{1}

\pgfmathsetmacro{\histogramstepwidth}{0.5}

\pgfdeclarelayer{gridlayer}    
\pgfdeclarelayer{background}    
\pgfsetlayers{gridlayer,background,main}  

\tikzset{pics/roadelement/.style ={code={
	\coordinate (-south west) at (-\elementwidth/2, 0);
	\coordinate (-south east) at ( \elementwidth/2, 0);
	\coordinate (-north west) at (-\elementwidth/2, \elementlength);
	\coordinate (-north east) at ( \elementwidth/2, \elementlength);
	\coordinate (-east)       at (\elementwidth/2, \elementlength / 2);
	\coordinate (-west)       at (-\elementwidth/2, \elementlength / 2);
	
	\coordinate (-reference)  at (0,0);
	\coordinate (-referencetop)  at (0,\elementlength);
	\draw[thick,  line join=round] (-south east) rectangle (-north west);
	\fill[tubsRed, draw=black] (0,0) circle (1.5pt);
	\pgfmathsetmacro{\histogramsteps}{\elementwidth / \histogramstepwidth}
	\pgfmathsetmacro{\histogramstepsminorone}{\elementwidth / \histogramstepwidth - 1}
	
	\pgfplotsset{/pgfplots/colormap name={fusion}}
	 
	\foreach \x in {1,2,...,\histogramsteps}
	{
	  		\draw[densely dashed, stealth'-, draw=tubsGray60] 
	  		   (-\elementwidth/2.0 + \x * \histogramstepwidth - \histogramstepwidth/2.0, 0.15) 
	  		-- (-\elementwidth/2.0 + \x * \histogramstepwidth - \histogramstepwidth / 2.0,\elementlength);
	}
	
	\foreach \color [count=\x] in {#1}
	{
		\pgfmathsetmacro{\colorindex}{\color * 1000}
		\draw[
			color of colormap={\colorindex},
			fill=.,
			draw=black,
			] (-\elementwidth/2.0 + \x * \histogramstepwidth -\histogramstepwidth*0.475 - \histogramstepwidth/2.0 , 0.05) rectangle ++(\histogramstepwidth*0.95,0.1);
    }
    }},
	pics/roadelement/.default={{}}
};

    \pgfmathsetmacro{\fc}{\freecolor}
    \pgfmathsetmacro{\uc}{\unknowncolor}
	\pgfmathsetmacro{\rc}{\markingcolor}
    \pgfmathsetmacro{\cc}{\curbcolor}
    \pgfmathsetmacro{\oc}{\obstaclecolor}
    
\begin{pgfonlayer}{gridlayer}    
	\pgfmathsetmacro{\gridsizex}{7}
	\pgfmathsetmacro{\gridsizey}{4}

    \draw (-3.5,-1) pic {gridlayout};
	\draw (-3.5,-1) pic[opacity=0.45] {drawgridcontents={fusion}%
		{\uc,\uc,\uc,\uc,\uc,\uc,\uc,\uc,\uc,\uc,\uc,\uc,\uc,\uc,\uc,\uc, 
         \uc,\uc,\uc,\uc,\uc,\uc,\uc,\uc,\uc,\uc,\uc,\uc,\uc,\uc,\uc,\uc, 
         \uc,\uc,\uc,\uc,\oc,\oc,\oc,\oc,\oc,\oc,\oc,\oc,\oc,\uc,\uc,\oc, 
         \oc,\oc,\oc,\oc,\uc,\oc,\oc,\oc,\oc,\uc,\oc,\oc,\oc,\oc,\oc,\oc, 
         \fc,\fc,\fc,\fc,\fc,\fc,\fc,\fc,\fc,\fc,\fc,\fc,\fc,\fc,\fc,\oc, 
         \fc,\fc,\fc,\fc,\fc,\fc,\fc,\fc,\fc,\fc,\fc,\fc,\fc,\fc,\fc,\fc, 
         \fc,\fc,\fc,\cc,\cc,\cc,\cc,\cc,\cc,\cc,\cc,\cc,\cc,\cc,\cc,\cc, 
         \fc,\fc,\fc,\fc,\fc,\fc,\fc,\cc,\cc,\fc,\fc,\fc,\fc,\fc,\fc,\fc, 
         \fc,\fc,\fc,\fc,\fc,\fc,\fc,\fc,\fc,\fc,\fc,\fc,\fc,\fc,\fc,\fc, 
         \fc,\fc,\fc,\fc,\fc,\fc,\fc,\fc,\fc,\fc,\fc,\fc,\fc,\fc,\fc,\fc, 
         \fc,\fc,\fc,\fc,\fc,\fc,\fc,\fc,\fc,\fc,\fc,\fc,\fc,\fc,\fc,\fc, 
         \fc,\fc,\fc,\fc,\fc,\fc,\fc,\fc,\fc,\fc,\fc,\fc,\fc,\fc,\fc,\fc, 
         \fc,\fc,\fc,\fc,\fc,\fc,\fc,\fc,\fc,\fc,\fc,\fc,\fc,\fc,\fc,\fc, 
         \fc,\fc,\fc,\fc,\fc,\fc,\fc,\fc,\fc,\fc,\fc,\fc,\fc,\fc,\fc,\fc, 
         \fc,\fc,\rc,\rc,\rc,\rc,\fc,\fc,\fc,\fc,\rc,\rc,\rc,\rc,\fc,\fc, 
         \fc,\fc,\fc,\fc,\fc,\fc,\fc,\fc,\fc,\fc,\fc,\fc,\fc,\fc,\fc,\fc, 
         \fc,\fc,\fc,\fc,\fc,\fc,\fc,\fc,\fc,\fc,\fc,\fc,\fc,\fc,\fc,\fc, 
         \fc,\fc,\fc,\fc,\fc,\fc,\fc,\fc,\fc,\fc,\fc,\fc,\fc,\fc,\fc,\fc, 
         \fc,\fc,\fc,\fc,\fc,\fc,\fc,\fc,\fc,\fc,\fc,\fc,\fc,\fc,\fc,\fc, 
         \fc,\fc,\fc,\fc,\fc,\fc,\fc,\fc,\fc,\fc,\fc,\fc,\fc,\fc,\fc,\fc, 
         \rc,\rc,\rc,\rc,\rc,\rc,\rc,\rc,\rc,\rc,\rc,\rc,\rc,\rc,\rc,\rc, 
         \fc,\fc,\fc,\fc,\fc,\fc,\fc,\fc,\fc,\fc,\fc,\fc,\fc,\fc,\fc,\fc, 
         \cc,\cc,\cc,\oc,\oc,\oc,\oc,\oc,\cc,\cc,\fc,\fc,\fc,\fc,\fc,\fc, 
         \uc,\uc,\oc,\oc,\oc,\oc,\oc,\oc,\oc,\uc,\uc,\uc,\uc,\uc,\uc,\uc, 
         \uc,\uc,\oc,\oc,\oc,\oc,\oc,\oc,\oc,\cc,\cc,\cc,\cc,\cc,\cc,\cc, 
         \uc,\uc,\uc,\uc,\uc,\uc,\uc,\uc,\uc,\uc,\uc,\uc,\uc,\uc,\uc,\uc, 
         \oc,\oc,\oc,\oc,\oc,\oc,\oc,\oc,\oc,\oc,\oc,\oc,\oc,\oc,\oc,\oc, 
         \oc,\oc,\oc,\oc,\oc,\oc,\oc,\oc,\oc,\oc,\oc,\oc,\oc,\oc,\oc,\oc 
        }};
\end{pgfonlayer}

\draw (0,-0.85)             pic[rotate=3] (r-0) {roadelement={\oc, \fc, \cc, \fc, \fc, \fc, \rc, \fc, \fc, \rc, \oc, \oc}};
\draw (r-0-referencetop)    pic[rotate=2] (r-1) {roadelement={\oc, \fc, \cc, \fc, \fc, \fc, \rc, \fc, \fc, \rc, \oc, \oc}};
\draw (r-1-referencetop)    pic[rotate=1] (r-2) {roadelement={\oc, \fc, \cc, \fc, \fc, \fc, \rc, \fc, \fc, \rc, \cc, \cc}};
\draw (r-2-referencetop)    pic[rotate=1, tubsGray40, dashed] (r-3) {roadelement={\oc, \oc, \cc, \fc, \fc, \fc, \rc, \fc, \fc, \rc, \fc, \cc}};

\draw[densely dotted, thick, draw=tubsGray60] (r-0-reference) -- ++ (1.5em, -1.5em) node[right, black] {$\vec{p}_i = \left(x_i, y_i, \varPsi_i \right)^T$};
 
\draw (r-0-east) node [right] {$RE_i$};
\draw (r-1-east) node [right] {$RE_{i+1}$};
\draw (r-2-east) node [right] {$RE_{i+2}$};
\draw (r-3-east) node [right,tubsGray60] {$RE_{i+n}$};
\draw[decorate, decoration={brace, amplitude=.5em, mirror, raise=0.2em}] (r-0-south west) -- (r-0-south east) node[midway, below, yshift=-0.6em] {$w$};
\draw[decorate, decoration={brace, amplitude=.5em, raise=0.2em}] (r-0-south west) -- (r-0-north west) node[midway, left, xshift=-0.8em] {$l$};

\begin{pgfonlayer}{background}    
	\draw[tubsGreenDark100, very thick, -stealth'] plot [smooth, tension=0.2] coordinates {(r-0-reference) (r-1-reference) (r-2-reference)(-0.1,3.4)} node[black, above, anchor=south west] {\raisebox{0pt}[1em][0pt]{Extraction path}};
\end{pgfonlayer}

\draw[decorate, decoration={brace, amplitude=0.5em, raise=0.2em}] ($(r-3-north west)!0.08!(r-3-referencetop)$) --($(r-3-north west)!0.92!(r-3-referencetop)$);
\node [above, anchor=south] at (-1.615, 3.4) {\raisebox{0pt}[1em][0pt]{Extraction lines}};

\end{tikzpicture}
    \tikzexternaldisable
    \vspace*{-1em}
    \caption[]{Generation of the interval-map representation according to \citet{Matthaei_Grid_Road_Detection_2013}. Given an extraction path, rectangular intervals of size $w \times l$ are defined (Road elements, REs), each one with an anchor~$\vec{p}_i$  along the path. The underlying grid cells are evaluated by accumulating their states along several extraction lines parallel to the interval direction. Colors correspond to different cell features, e.\,g. freespace (green), lane markings (beige), obstacle types (yellow, orange), or unknown regions (gray).}
    \label{fig:Grids_Roadelements}
    \vspace*{-1em}
\end{figure}

Based on the combined results of the fusion layer, different extractors can be applied to extract features for the context modeling stage.
This may be polyline-based representations of free and occupied regions \citep{Kubertschak_Fences_2014}, or spline-based representations \citep{Schreier_Parametric_Freespace_Maps_2013}. 
\citet{Matthaei_Grid_Road_Detection_2013} use an interval-based representation, based on the work of \citet{Weiherer_Interval_Maps_2012}, to extract road course information and perform lane-level localization \citep{Matthaei_Lane_Level_Localization_2014}.
We use Matthaei's work on interval-map structures to extract borders of driving corridors as well as elevated obstacles on and next to the road. 
As illustrated in  \autoref{fig:Grids_Roadelements}, the interval map approach constitutes a compact representation of different features in the vicinity of a given extraction path.
This path is usually corresponding to a potential driving corridor, thus the interval map describes elements that limit this corridor or are located next to it.

\section{Movable Environment Modeling}
\label{sec:Object_Tracking}

\providecommand{\currentimagedirectory}{}
\renewcommand{\currentimagedirectory}{./figures/6_Object_Tracking}

\begin{figure*}
    \centering
    \subfloat[Fusion grid layer]{\label{subfig:Results_Fusion_Grid}%
        \includegraphics[width=0.32\textwidth]{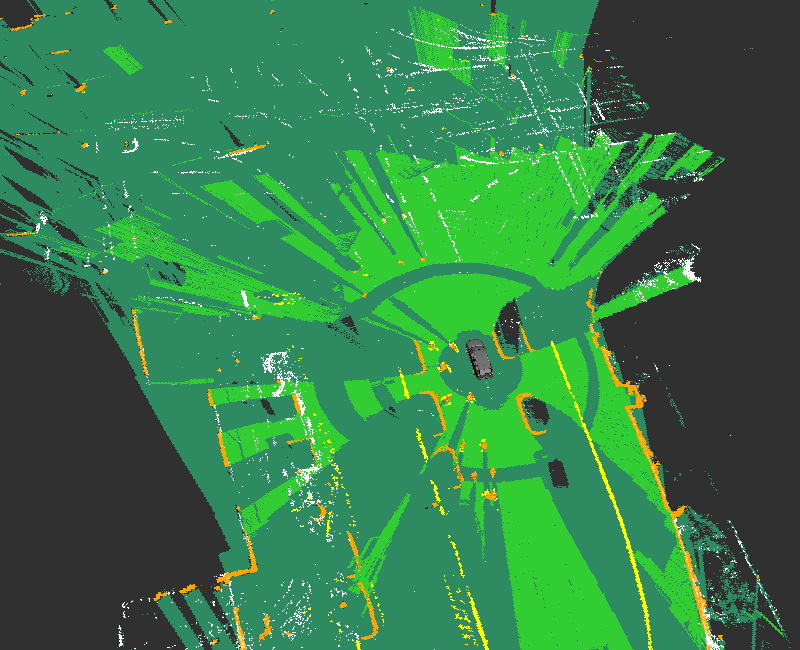}%
    } \hfill 
    \subfloat[Movable object hypotheses]{\label{subfig:Results_Hypotheses}%
        \includegraphics[width=0.32\textwidth]{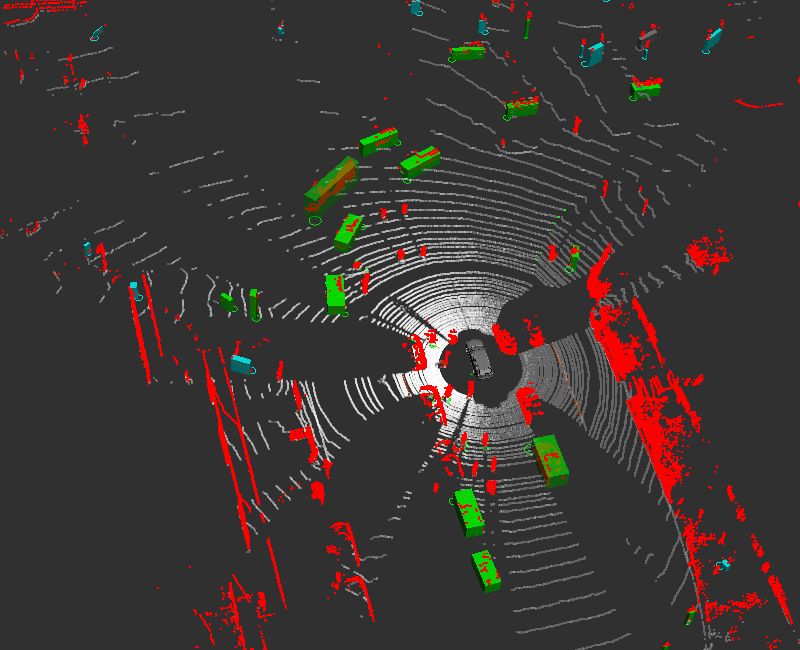}%
    }\hfill
    \subfloat[Stationary and movable objects]{\label{subfig:Results_Tracking}%
        \includegraphics[width=0.32\textwidth]{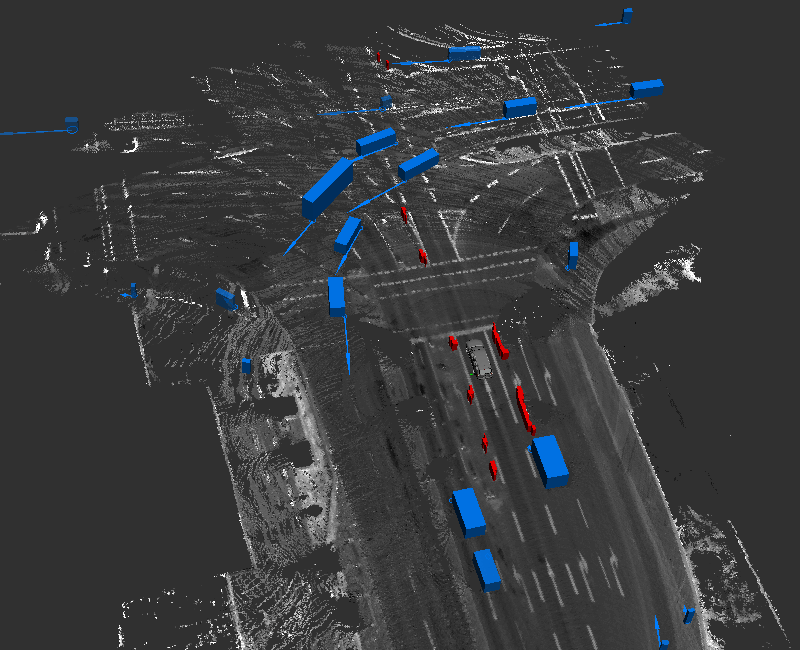}%
    }
    \caption{
        Results of the stationary and movable environment model.  
        \protect\subref{subfig:Results_Fusion_Grid} shows the fusion grid layer. Elevated targets are colored in orange, curb structures in yellow. High-reflectivity regions are painted white. The different shades of green illustrate implicit freespace information (from Bayesian Occupancy Grids, dark green) and explicit freespace information inferred from ground measurements (light green).
        \protect\subref{subfig:Results_Hypotheses} shows the object hypotheses for the current time instant (green: movable, cyan: potentially movable) together with the ground-classified 3D pointcloud.    
        The results from the object tracking module (blue boxes) are shown in \protect\subref{subfig:Results_Tracking}. 
        The image also contains the extracted stationary obstacles from the fusion layer (red cubes), which originate from safety beacons on the road in this scene.
    }
    \vspace{-1em}
\end{figure*}

The modeling of the movable environment is accomplished by an object-based approach using multi-instance Kalman Filters.
Herein, each movable element is represented by a 3D bounding box instance, denoted as \emph{object} in the following.
The bounding box model was chosen as a trade-off between tracking objects as point-like targets versus using shape-based approaches (see \citep{Held_Robust_Tracking_2014, Moosmann_Dissertation_2013}).
As our tracking system is designed to deal with nearby objects as well as those at a larger distance, a bounding box is suitable for representing the major properties of an object while it still allows to handle changes of the target's dimensions, such as due to occlusion and viewpoint changes, appropriately.
In the following, we explain the generation of box hypotheses from clustered pointclouds and our approach of tracking these hypotheses over time.
Results of both modules are shown in \autoref{subfig:Results_Hypotheses} and~\ref{subfig:Results_Tracking}.

\subsection{Hypotheses generation and object model}
The movable environment is generated from all non-stationary clusters, i.\,e., all clusters that do not belong to the class \emph{stationary}.
Each movable element is represented by an \emph{object model}.
It represents the outer contour of an element by its 3D bounding box.
Hypotheses are generated by a multi-step algorithm.
Here we include a brief overview of the algorithm while the interested reader is referred to \citep{Rieken_Environment_Modelling_2015} for more details.

From a top-view perspective, a convex polyline boundary is calculated for each cluster.
It is then classified as an I-, L- or trapezoid shape.
 Afterwards, an oriented bounding box is defined based on the shape's type and adjusted to include all measurements of the cluster.
Finally, the box height is determined from the measurements' vertical positions relative to the underlying ground surface.

\begin{figure}
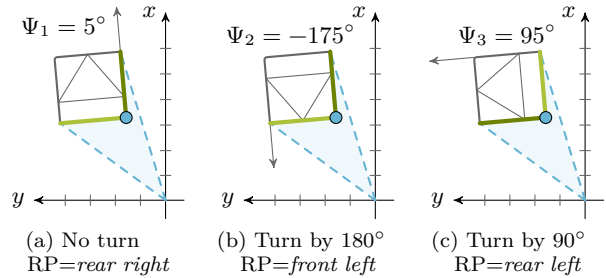

  	\centering
      \captionsetup[subfigure]{justification=centering}
  \subfloat[No turn \protect\newline RP=\protect\emph{rear right}]{\label{subfig:Systemrealisierung_Objektverfolgung_Assoziation_Referenzpunktmehrdeutigkeit_Fall_1}
      \begin{tikzpicture}[rotate=90, every node/.style={font={\small}}]

\input{\currentimagedirectory/objektverfolgung_pics_definition.tikz}
\pgfmathsetmacro{\objectlength}{\objectwidth}

\tikzstyle{commonobject}=[scale=0.4, rotate=90, objectcolor, rotate=5];

\draw[coordinateaxis] (-0.25, 0) -- (2.5,0) node[left] {$x$};
\foreach \x in {0.33,0.66,1,...,2.2}
{
	\draw[tubsGray60] (\x, -0.075) -- (\x, 0.0750);
}
\draw[coordinateaxis] (0, -0.25) -- (0,1.75) node[left] {$y$};
\foreach \y in {0.334,0.667,...,1.5}
{
	\draw[tubsGray60] (-0.075, \y) -- (0.075, \y);
}

\draw pic[commonobject] (object) at (1.5,1) {object};
\draw pic[commonobject] (objectdirection) at (object-FR) {objectorientationoverlay};

\fill[rayarea] (0,0) -- (object-RL) -- (object-RR) -- (object-FR);
\draw[ray] (0,0) -- (object-RL);
\draw[ray] (0,0) -- (object-FR);


\draw[objectsidewidth] (object-RL) -- (object-RR);
\draw[objectsidelength] (object-RR) -- (object-FR);

\draw pic[scale=0.75] at (object-RR) {referencepoint};
\draw (objectdirection-head) node [anchor=north east, xshift=0em] {$\Psi_1 = \ang{5}$};

\end{tikzpicture}
  }
  \subfloat[Turn by \ang{180} \protect\newline RP=\protect\emph{front left}]{\label{subfig:Systemrealisierung_Objektverfolgung_Assoziation_Referenzpunktmehrdeutigkeit_Fall_2}
      \begin{tikzpicture}[rotate=90, every node/.style={font={\small}}]

\input{\currentimagedirectory/objektverfolgung_pics_definition.tikz}
\pgfmathsetmacro{\objectlength}{\objectwidth}
\tikzstyle{commonobject}=[scale=0.4, rotate=90, objectcolor, rotate=185];

\draw[coordinateaxis] (-0.25, 0) -- (2.5,0) node[left] {$x$};
\foreach \x in {0.33,0.66,1,...,2.2}
{
	\draw[tubsGray60] (\x, -0.075) -- (\x, 0.0750);
}
\draw[coordinateaxis] (0, -0.25) -- (0,1.75) node[left] {$y$};
\foreach \y in {0.334,0.667,...,1.5}
{
	\draw[tubsGray60] (-0.075, \y) -- (0.075, \y);
}

\draw pic[commonobject] (object) at (1.5,1) {object};
\draw pic[commonobject] (objectdirection) at (object-FR) {objectorientationoverlay};

\fill[rayarea] (0,0) -- (object-FR) -- (object-FL) -- (object-RL);
\draw[ray] (0,0) -- (object-RL);
\draw[ray] (0,0) -- (object-FR);


\draw[objectsidewidth] (object-FR) -- (object-FL);
\draw[objectsidelength] (object-RL) -- (object-FL);

\draw pic[scale=0.75] at (object-FL) {referencepoint};
\draw (object-RC) node [above, xshift=-0.25em ] {$\Psi_2 = \ang{-175}$};

\end{tikzpicture}
  }
  \subfloat[Turn by \ang{90} \protect\newline RP=\protect\emph{rear left}]{\label{subfig:Systemrealisierung_Objektverfolgung_Assoziation_Referenzpunktmehrdeutigkeit_Fall_3}
      \begin{tikzpicture}[rotate=90, every node/.style={font={\small}}]

\input{\currentimagedirectory/objektverfolgung_pics_definition.tikz}
\pgfmathsetmacro{\objectlength}{\objectwidth}

\tikzstyle{commonobject}=[scale=0.4, rotate=90, objectcolor, rotate=95];

\draw[coordinateaxis] (-0.25, 0) -- (2.5,0) node[left] {$x$};
\foreach \x in {0.33,0.66,1,...,2.2}
{
	\draw[tubsGray60] (\x, -0.075) -- (\x, 0.0750);
}
\draw[coordinateaxis] (0, -0.25) -- (0,1.75) node[left] {$y$};
\foreach \y in {0.334,0.667,...,1.5}
{
	\draw[tubsGray60] (-0.075, \y) -- (0.075, \y);
}

\draw pic[commonobject] (object) at (1.5,1) {object};
\draw pic[commonobject] (objectdirection) at (object-FR) {objectorientationoverlay};

\fill[rayarea] (0,0) -- (object-FL) -- (object-RL) -- (object-RR);
\draw[ray] (0,0) -- (object-RR);
\draw[ray] (0,0) -- (object-FL);

\draw[objectsidewidth] (object-RR) -- (object-RL);
\draw[objectsidelength] (object-FL) -- (object-RL);

\draw pic[scale=0.75] at (object-RL) {referencepoint};
\draw (object-MR) node [above] {$\Psi_3 = \ang{95}$};

\end{tikzpicture}
  }
    \caption{Ambiguities of the bounding box representation regarding the reference point (RP) label of the selected corner. Here, the nearest corner is selected as reference point (blue circle). Shown are three of four possible interpretations of the resulting object orientation. 
    For~\ang{\pm 90} turns of the box orientation~$\Psi$, length and width (dark resp. light green sides) of the bounding box get exchanged.}
    \label{fig:Ambiguity_Referencepoint}
    \vspace*{-1em}
\end{figure}

In addition to each hypothesis' volumetric extent and its orientation in the $x$-$y$-plane, information about the reference point is added. 
We use the so-called \emph{best-knowledge} model published by \citet{Schueler_Sensor_Fusion_2012}.
Its idea is to represent each object by the most reliable reference point, i.\,e. the point which is the best observed one.
This model defines a total of nine reference points for an oriented bounding box, four of which are located on the corners of the rectangle, four on the sides and one in the center. 
Using this reference point model avoids undesired artificial velocities, as we can compensate for occlusion-related changes in the shapes of the detected objects.
The model explicitly distinguishes between the corner and sides of a target by giving them semantic tags.
LiDAR-based sensors are hardly able to classify the seen portions of an element, which makes the assignment of a reference point ambiguous regarding to e.\,g. the detected edge, as illustrated in \autoref{fig:Ambiguity_Referencepoint}.   
To handle this explicitly, we extend this model by a notation of the ambiguity of the chosen reference point.

\subsection{Occlusion reasoning}
\label{ssec:Tracking_Occlusion_reasoning}
\begin{figure}
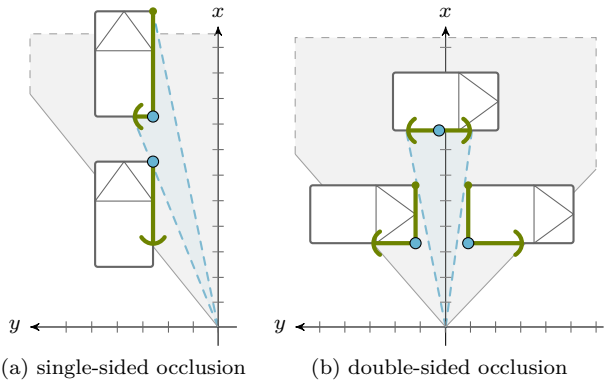

    \centering
    \subfloat[single-sided occlusion]{
        \begin{tikzpicture}[rotate=90]

\input{\currentimagedirectory/objektverfolgung_pics_definition.tikz}

\pgfdeclarelayer{background}
\pgfsetlayers{background,main}

\draw[coordinateaxis] (-0.25, 0) -- (4,0) node[above] {$x$};
\foreach \x in {0.33,0.66,1,...,3.9}
{
	\draw[tubsGray60] (\x, -0.075) -- (\x, 0.0750);
}
\draw[coordinateaxis] (0, -0.25) -- (0,2.5) node[left] {$y$};
\foreach \y in {-0,0.33,0.66,1,...,2.3}
{
	\draw[tubsGray60] (-0.075, \y) -- (0.075, \y);
}

\coordinate (fovleft) at (3.1, 2.5);
\draw[sensorfieldofview] (0,0) -- (fovleft);
\draw[sensorfieldofview_outer_border] (fovleft) -- (3.9, 2.5) -- (3.9, 0);
\fill [tubsGray20, opacity=0.25] (0,0) -- (fovleft) -- (3.9, 2.5) -- (3.9, 0) -- cycle;
\draw pic[scale=0.35, rotate=90, objectcolor] (preceedingobject) at (3.5,1.25) {object};
\draw pic[scale=0.35, rotate=90, objectcolor] (occludingobject) at (1.5,1.25) {object};

\draw[visiblecontourline, (-] ($(preceedingobject-RL)!0.65!(preceedingobject-RR)$) coordinate (preceedingobject-fovlimit) -- (preceedingobject-RR);
\draw[visiblecontourline, -](preceedingobject-RR) -- (preceedingobject-FR);
\fill[visiblecontourline] (preceedingobject-FR) circle(1.5pt);


\draw[visiblecontourline, (-] ($(occludingobject-RR)!0.20!(occludingobject-FR)$) coordinate (occludedobject-fovlimit) -- (occludingobject-FR);

\begin{pgfonlayer}{background}
	\fill[rayarea] (0,0) -- (preceedingobject-FR) -- (preceedingobject-RR) -- (preceedingobject-fovlimit);
	\draw[ray] (0,0) -- (preceedingobject-FR);

	\draw[ray] (0,0) -- (preceedingobject-fovlimit);
\end{pgfonlayer}

\draw pic[scale=0.7] at (preceedingobject-RR) {referencepoint};
\draw pic[scale=0.7] at (occludingobject-FR) {referencepoint};

\end{tikzpicture}
    }
    \subfloat[double-sided occlusion]{
        \begin{tikzpicture}[rotate=90]

\input{\currentimagedirectory/objektverfolgung_pics_definition.tikz}

\pgfdeclarelayer{background}
\pgfsetlayers{background,main}

\draw[coordinateaxis] (-0.25, 0) -- (4,0) node[above] {$x$};
\foreach \x in {0.33,0.66,1,...,3.9}
{
	\draw[tubsGray60] (\x, -0.075) -- (\x, 0.0750);
}
\draw[coordinateaxis] (0, -2.1) -- (0,2) node[left] {$y$};
\foreach \y in {-2,-1.666,...,2}
{
	\draw[tubsGray60] (-0.075, \y) -- (0.075, \y);
}

\coordinate (fovleft) at (2.3, 2);
\coordinate (fovright) at (2.1, -2);

\draw[sensorfieldofview] (0,0) -- (fovleft);
\draw[sensorfieldofview] (0,0) -- (fovright);

\draw[sensorfieldofview_outer_border] (fovleft) -- (3.85,2) -- (3.85,-2) -- (fovright);
\fill [tubsGray20, opacity=0.25] (0,0) -- (fovleft) -- (3.85,2) -- (3.85,-2) -- (fovright) -- cycle;

\draw pic[scale=0.35, rotate=0, objectcolor] (occludingobjectleft)  at (1.5, 1.1)  {object};
\draw pic[scale=0.35, rotate=0, objectcolor] (occludingobjectright) at (1.5,-1.0)  {object};
\draw pic[scale=0.35, rotate=0, objectcolor] (relevantobject)       at (3,0)     {object};

\coordinate (relevantobjectfovleft) at ($(relevantobject-RR)!0.125!(relevantobject-FR)$);
\coordinate (relevantobjectfovright) at ($(relevantobject-RR)!0.75!(relevantobject-FR)$);
\coordinate (relevantobjectcenter) at ($(relevantobjectfovleft)!0.5!(relevantobjectfovright)$);
\draw[visiblecontourline, (-)] (relevantobjectfovleft) -- (relevantobjectfovright);

\coordinate (occludingobjectleftfovleft) at  ($(occludingobjectleft-RR)!0.58!(occludingobjectleft-FR)$);
\coordinate (occludingobjectleftfovright) at (occludingobjectleft-FL);
\draw[visiblecontourline, (-] (occludingobjectleftfovleft) -- (occludingobjectleft-FR) -- (occludingobjectleftfovright);
\fill[contourline] (occludingobjectleftfovright) circle(1.5pt);

\coordinate (occludingobjectrightfovleft) at  (occludingobjectright-RL);
\fill[contourline] (occludingobjectrightfovleft) circle(1.5pt);
\coordinate (occludingobjectrightfovright) at ($(occludingobjectright-RR)!0.53!(occludingobjectright-FR)$);
\draw[visiblecontourline, -)] (occludingobjectrightfovleft) -- (occludingobjectright-RR) -- (occludingobjectrightfovright);

\begin{pgfonlayer}{background}
	\fill[rayarea] (0,0) -- (relevantobjectfovleft) -- (relevantobjectfovright);
	\draw[ray] (0,0) -- (relevantobjectfovleft);
	\draw[ray] (0,0) -- (relevantobjectfovright);
\end{pgfonlayer}

\draw pic[scale=0.7] at (occludingobjectright-RR) {referencepoint};
\draw pic[scale=0.7] at (occludingobjectleft-FR) {referencepoint};
\draw pic[scale=0.7] at (relevantobjectcenter) {referencepoint};

\end{tikzpicture}
    }

    \caption[]{Illustration of single- and double-sided occlusion due to a limited field of view or occlusion by another object. Half-circled arrowheads (\begin{tikzpicture}[baseline=-0.25em]\draw[-), thick, tubsGreenMedium100] (0,0)--(0.25,0);  \end{tikzpicture}) mark the visible proportions of the occluded sides of the bounding box.
    The resulting reference point is drawn by a blue circle. The sensor's field of view is indicated in shaded gray.}
    \label{fig:Evidential_classes}
    \vspace*{-1em}
\end{figure}

The hypotheses generation step does not account for occlusions due to other movable or stationary elements, but chooses the nearest edge as reference point. 
Yet, in case of occlusion, this might not be an actual edge, but an artifact due to the occlusion (see \autoref{fig:Evidential_classes}).
We introduce a ray-tracing approach to detect such constellations.
Due to the availability of a channel-based sensor data structure, this becomes a simple check for targets at smaller distances in those channels that correspond to the object's edges.

Based on those results, a quality class~$\mathcal{E}_f$ is assigned to each dimensional state \mbox{$f \in \{\text{length}, \text{width}, \text{height} \}$}.
The class describes the context in which each dimensional state has been retrieved (see \autoref{fig:Evidential_classes}):
\begin{description}
    \item[bounded] The dimension was inferred from the detection of both limiting edges.
    \item[single-sided open interval] The dimension was measured by detecting one limiting edge, which is marked by the reference point. The full length of the side could not be obtained, but corresponds to the currently visible part of the object.
    \item[double-sided open interval] The state was derived from two non-limiting edges, e.\,g. due to occlusions on both sides. The reference point marks the center of the dimension, but might not correspond to the real side's center.
\end{description}

\subsection{Object tracking}
\label{ssec:Object_tracking}
The object hypotheses are filtered and validated using an Extended Kalman Filter (EKF).
Based on a greedy object classifier, which derives a class estimation from the objects dimensions and velocity ranges, a finite state machine switches between a simple constant velocity (CV) motion model for unclassified targets and an Interacting Multiple Model (IMM) approach for vehicle-class objects.
The IMM filter is set up with two motion models, namely a constant acceleration (CA) and a constant turn-rate and velocity (CTRV) model.
In the following, state prediction and update steps are described in a general fashion without explicitly addressing additional calculations required by the IMM model.
Objects' states reflect their poses and motion in the vehicle reference frame, but in terms of an ego-independent motion. 
Thus, the state prediction step also considers the ego motion within the relevant time slices.  

Based on previous results (\citep{Rieken_Environment_Modelling_2015, Rieken_Scan_Timing_2016}), the following section focuses on the integration of the aforementioned object model extensions. 
The results will be explained along the basic processing steps of recursive state estimation techniques.
\subsubsection{State prediction and association}
Each object is represented by a state vector, whose dimensions depend on the currently assigned motion model.
The motion model predicts the objects' states to the timestamps of the incoming hypotheses.
For rotating sensors like scanning LiDARs, timestamps differ for each detected object. 
Thus it is not reasonable to predict all objects by the same time step~$\Delta T$, but the prediction time step has to be determined for each object separately.
We incorporate a runtime-efficient approximative calculation of this time interval, based on a sensor timing model \citep{Rieken_Scan_Timing_2016}.

Association between new hypotheses and already known objects uses a Local Nearest Neighbor approach.
The similarity of each assignment pair is evaluated in terms of the Mahalanobis distance. 
As different reference points of object and hypothesis typically introduce additional distance errors, we transform the object's state to the same reference point label as assigned to the hypothesis.
Distance calculation is performed in the hypothesis' state space. 
To account for erroneous object extents caused by occlusions, the dimensional states of both object and hypothesis are incorporated in the distance function depending on their quality classes~$\mathcal{E}_f$:
If a partially observed state (i.\,e., a state with a quality class other than \emph{bounded}) indicates a smaller object extent than currently estimated by the tracker, the respective dimension is ignored for association.
Depending on the annotated reference point ambiguity, multiple interpretations of each hypothesis are considered. 
Up to now, reference point ambiguity is resolved by selecting the interpretation that results in the smallest orientation error for the association distance.

\subsubsection{State update}
Each existing object with an associated hypothesis is updated using the (IMM-)EKF equations.
Like in the association step, the considered states depend on the quality classes of the dimensional states of object and hypothesis.
Potential ambiguity of the reference point is resolved by evaluating the estimated motion direction of the object.

\subsubsection{Object lifecycle}
New objects are instantiated from all unassociated hypotheses.
Objects are considered tentative until they have been updated for at least three times and have moved at least~\SI{2}{\meter} from their point of first detection.
They are published to downstream modules only after having fulfilled both conditions.
Objects are deleted from the tracker's database if they exceed their \emph{time-to-live}.
This value describes the object's plausibility by the time difference since the last state update.

\section{Evaluation}
\label{sec:Evaluation}

\providecommand{\currentimagedirectory}{}
\renewcommand{\currentimagedirectory}{./figures/7_Evaluation}

The following section presents the evaluation of the presented perception system.
We perform the evaluation at the interface to the context/scene modeling module (see \autoref{fig:Sytem_Design_Architecture}) and focus on the movable environment part. 
In order to achieve more transferable results regarding the system's real-world performance, we propose to evaluate the system in the context of its actual driving task.

For the scope of this paper, we select three scenarios that automated vehicles have to cope with\footnote{
    Following the definition of \citet{Ulbrich_Scene_Scenario_2015}, each scenario is defined by a time series of snapshot-like scenes, along with goals and values of each participant, especially that ones of the \emph{own} vehicle (cf. \citet[Sec.~VI]{Ulbrich_Scene_Scenario_2015}). Hence, the definition of each scenario also contains the description of the automated vehicle's mission.}.
These scenarios were chosen by the authors to represent common and rather challenging driving tasks and traffic constellations, specifically along the intended route of the project \emph{Stadtpilot}.
Please note that this list does not impose any claim for completeness or a representative selection of test scenarios for urban automated driving as such.
For each scenario we recorded real-world raw data of public traffic from test drives along the project's route. An object-level \emph{ground-truth} annotation of the movable traffic participants was created in a post-processing step.

\subsection{Open-loop evaluation strategies and their limitations}
The most predominate approach to evaluate object-based data representations is to compare the system's output with ground-truth data sets.
These data sets are assumed to represent the real environment in perfection;
in general, the evaluated system is expected to show comparably inferior performance.
By applying a set of evaluation criteria, results such as detection rates and error distributions can be obtained.
As this approach does not consider any actual self-driving functions by default, we refer to it as \emph{open-loop} evaluation.

 Due to this missing link to the system's intended usecases and its operational design domain (\emph{ODD}, see \citet{SAE_J3016}), an open-loop evaluation may not able to reflect the \emph{criticality} of errors in the context of the actual driving task.
     As an example, not detecting a closeby pedestrian entering the road is much more severe than missing an object at a larger distance, from the perspective of an emergency braking system.
    However, typical error metrics will treat both cases equally, which is misleading for a performance assessment of the system.

\subsection{Closed-loop approach}
To overcome this constraint, we propose to evaluate the perception system in the context of the current driving task of the automated vehicle.
As this includes perception as well as behavior planning components, we refer to this as \emph{closed-loop} evaluation.

In terms of the functional system architecture of \citet{Matthaei_FSA_2015}, driving behavior generation is preceded by a \emph{situation assessment}.
As an initial step, a tactical planning module abstracts and augments the provided environment representation (\emph{scene model}, see \autoref{fig:Sytem_Design_Architecture}) to a more focused structure.
The current mission of the vehicle and its possible tactical decisions to reach its mission's goal are considered here.
Hence, this step also defines the \emph{relevancy} of different parts of the environment to the driving function, among other items.
This \emph{augmented scene description} is a central component for the evaluation of the perception, since it creates the link between the perception results and the driving function.

In the following, we first infer a \emph{ground-truth} augmented scene description by using the ground-truth object list as input.
This step is then repeated with the object list generated by the presented perception system.
Comparing both augmented scenes enables us to focus on the \emph{relevant} elements for the driving function and to evaluate whether the perception system is able to represent these elements appropriately.

Since the evaluation is performed on pre-recorded data, actions decided for during the recording, along with possible reactions from other traffic participants, cannot be altered.
Using such type of data requires a \emph{stateless} behavior planning module, which infers decisions based on the current frame only.
However, for comparably simple tasks as described in the following sections, this type of planning module can be assumed.

\subsection{Definition of usecases and selected scenarios}
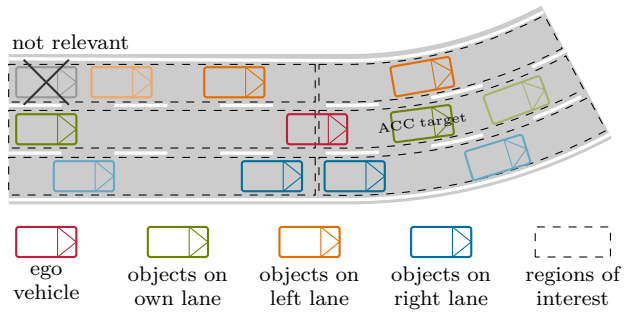
\begin{figure}
    \centering
        \begin{tikzpicture}[every node/.style={font=\footnotesize}, scale=0.5]

	\tikzstyle{road}=[fill=tubsGray20, draw=none]
	\tikzstyle{marking}=[fill=none, draw=white, line width=1.25]
	\tikzstyle{dashedmarking}=[marking, dash pattern=on 20 off 20]
	\tikzstyle{ROImarking}=[fill=none, tubsBlack, dashed]

	\pgfmathsetmacro{\roadwidth}{4}
	\pgfmathsetmacro{\roadlength}{9}
	\pgfmathsetmacro{\markingwidth}{0.05}
	\pgfmathsetmacro{\markingborderdistance}{0.125}
	\pgfmathsetmacro{\startangle}{-90}
	\pgfmathsetmacro{\endangle}{-62}
	\pgfmathsetmacro{\roadradius}{13}
	\pgfmathsetmacro{\roadradiusinner}{\roadradius - \roadwidth/2}
	\pgfmathsetmacro{\roadradiusouter}{\roadradius + \roadwidth/2}
	\pgfmathsetmacro{\markingoneradius}{\roadradius - \roadwidth/6.3}	
	\pgfmathsetmacro{\markingtworadius}{\roadradius + \roadwidth/6.3}	
	\pgfmathsetmacro{\outermarkingradius}{\roadradiusouter - \markingborderdistance}	
	\pgfmathsetmacro{\innermarkingradius}{\roadradiusinner + \markingborderdistance}
	\pgfmathsetmacro{\legendrowy}{-3}	
	\pgfmathsetmacro{\legendrowx}{0}
	\pgfmathsetmacro{\legendrowstepx}{4}
	\pgfmathsetmacro{\lanewidth}{1.0}
	
	\pgfmathsetmacro{\objectlength}{\roadwidth / 5}
	\pgfmathsetmacro{\objectwidth}{\roadwidth / 10}

\tikzset{pics/object/.style={code={
	\pgfmathsetmacro{\objectcenterx}{\objectwidth/2}
	\pgfmathsetmacro{\objectcentery}{\objectlength/2}
	
	\draw[rounded corners=1pt, thick] (-\objectcentery,-\objectcenterx) rectangle ++(\objectlength, \objectwidth);
	\draw[line cap=round] (\objectcentery - 0.25 , -\objectcenterx) -- (\objectcentery - 0.25, \objectcenterx) -- (\objectcentery,0) -- cycle;
	
	\coordinate (-FL) at (+\objectcentery, +\objectcenterx)  ;
	\coordinate (-MC) at (0,0) 							     ;
	\coordinate (-RL) at (-\objectcentery, +\objectcenterx)  ;
	\coordinate (-RR) at (-\objectcentery, -\objectcenterx)  ;
	\coordinate (-FR) at (+\objectcentery, -\objectcenterx)  ;	                                                      
	\coordinate (-RC) at (-\objectcentery, 0)				 ;
	\coordinate (-FC) at (+\objectcentery, 0)				 ;
	\coordinate (-MR) at (0, -\objectcenterx)				 ;
	\coordinate (-ML) at (0, +\objectcenterx)				 ;
    \coordinate (-Tail) at (-0.5 * \objectcentery + 0.375, 0);
}}};

	\path (\roadlength, \roadwidth/2) arc[start angle=\startangle, end angle=\endangle, radius=\roadradiusinner] coordinate (laneend) {};
	
	\path [road]
	 (0, -\roadwidth/2) -- (\roadlength, -\roadwidth /2) arc[start angle=\startangle, end angle=\endangle, radius=\roadradiusouter]
	 --(laneend)
	 arc[start angle=\endangle, end angle=\startangle, radius=\roadradiusinner]
	 --(0, \roadwidth/2) -- (0, -\roadwidth /2);
	  
	\draw[marking] (0,-\roadwidth /2 + \markingborderdistance) -- (\roadlength, -\roadwidth /2 + \markingborderdistance) 
	arc[start angle=\startangle, end angle=\endangle, radius=\outermarkingradius];
	
	\draw[marking] (0,+\roadwidth /2 - \markingborderdistance) -- ++(\roadlength, -\markingwidth) arc[start angle=\startangle, end angle=\endangle, radius=\innermarkingradius];
	
	\draw[dashedmarking] (0,+\roadwidth /6.3) -- ++(\roadlength, 0)  arc[start angle=\startangle, end angle=\endangle, radius=\markingoneradius];
	\draw[dashedmarking] (0,-\roadwidth / 6.3) -- ++ (\roadlength, 0) arc[start angle=\startangle, end angle=\endangle, radius=\markingtworadius];

	\draw[ROImarking] (8.15, \lanewidth * 0.5) rectangle ++(-8.15, -\lanewidth);
	\draw[ROImarking] (8.15, \lanewidth * 1.5 + 0.225) rectangle ++(-8.15, -\lanewidth);
	\draw[ROImarking] (8.15, -\lanewidth * 0.5 - 0.25) rectangle ++(-8.15, -\lanewidth);

	\path[tubsRed] (\roadlength, -\lanewidth * 0.5  ) arc[start angle=\startangle, end angle=\endangle, radius=\roadradius + \roadwidth/6.5 -0.225] coordinate (laneend) {};
	
	\draw[ROImarking] (8.25, \lanewidth*0.5) -- (\roadlength, \lanewidth*0.5) arc[start angle=\startangle, end angle=\endangle, radius=\roadradius - \roadwidth/6.5] -- (laneend)  arc[start angle=\endangle, end angle=\startangle, radius=\roadradius + \roadwidth/6.5 -0.20] -- (8.25, -\lanewidth * 0.5) -- cycle;

	\path[tubsRed] (\roadlength, \lanewidth * 0.5+0.225) arc[start angle=\startangle, end angle=\endangle, radius=\roadradius + \roadwidth/6.5-1.45] coordinate (laneend) {};
	
	\draw[ROImarking] (8.25, \lanewidth * 1.5 +0.225) -- (\roadlength, \lanewidth * 1.5  +0.225) arc[start angle=\startangle, end angle=\endangle, radius=\roadradius - \lanewidth/2 - \roadwidth/3] -- (laneend) arc[start angle=\endangle, end angle=\startangle, radius=\roadradius + \roadwidth/6.5-1.45] -- (8.25, \lanewidth * 0.5+0.225) -- cycle;
	
	\path[tubsRed] (\roadlength, -\lanewidth *1.5 - 0.25) arc[start angle=\startangle, end angle=\endangle, radius=\roadradius + \roadwidth/6.5+1.05] coordinate (laneend) {};
	
	\draw[ROImarking] (8.25, -\lanewidth * 0.5 - 0.25) -- (\roadlength, -\lanewidth * 0.5 - 0.25) arc[start angle=\startangle, end angle=\endangle, radius=\roadradius + \lanewidth/2 +0.15]-- (laneend) arc[start angle=\endangle, end angle=\startangle, radius=\roadradius + \roadwidth/6.5+1.15] -- (8.25, -\lanewidth * 1.5 - 0.25) -- cycle;

	\draw (8.2,0) pic[tubsRed] (ego) {object};
	\draw (6,1.25) pic[tubsOrangeMedium100] (object-HL) {object};
	\draw (3,1.25) pic[tubsOrangeMedium60] (object-HL2) {object};
	\draw (7,-1.25) pic[tubsBlueMedium100] (object-HR) {object};
	\draw (2,-1.25) pic[tubsBlueMedium60] (object-HR2) {object};
	\draw (1, 1.25) pic[tubsGray40] (object-invalid) {object};
	
	\draw (11,1.4) pic[tubsOrangeMedium100, rotate=10] (object-FL) {object};
	\draw (9.2,-1.25) pic[tubsBlueMedium100] (object-FR) {object};
	\draw (13,-0.75) pic[tubsBlueMedium60, rotate=17] (object-FR2) {object};
	
	\draw (11,0.15) pic[tubsGreenMedium100,rotate=7.5] (object-FC) {object};
	\draw (13.5,0.78)  pic[tubsGreenMedium60, rotate=20] (object-FC2) {object};
	\draw (1,0)     pic[tubsGreenMedium100] (object-HC2) {object};
	
	\node[rotate=7.5, font={\tiny}] at (object-FC-MC) {ACC target};

	\draw[tubsGray80, thick] (object-invalid-MC) ++ (-0.6, 0.6) -- ++ (1.2,-1.2);
	\draw[tubsGray80, thick] (object-invalid-MC) ++ (0.6, 0.6) -- ++ (-1.2,-1.2);
	\path (object-invalid-MC) ++ (0.65,0.65) node [above] {not relevant};
	\draw (1., \legendrowy) pic[tubsRed] (object-Center-L) {object};
	\draw (object-Center-L-MR) node[below,align=center] {ego \\ vehicle};
	\draw (4.5, \legendrowy) pic[tubsGreenMedium100] (object-Center-L) {object};
	\draw (object-Center-L-MR) node[below,align=center] {objects on \\ own lane};
	\draw (8, \legendrowy) pic[tubsOrangeMedium100] (object-Center-L) {object};
	\draw (object-Center-L-MR) node[below,align=center] {objects on \\ left lane};
	\draw (11.5, \legendrowy) pic[tubsBlueMedium100] (object-Center-L) {object};
	\draw (object-Center-L-MR) node[below,align=center] {objects on\\right lane};
	\draw[ROImarking] (15, \legendrowy) ++ (-\roadwidth / 4, -\roadwidth / 10) rectangle ++(\roadwidth / 2, \roadwidth / 5);
	\draw (15, \legendrowy) ++ (0, -\roadwidth / 10) node[below,align=center] {regions of\\interest};
	
\end{tikzpicture}
    \vspace*{-1em}
    \caption{Augmented scene description for lane change decision planning by \citet{Ulbrich_POMDP_2013}, which is utilized here for the definition of regions of interest for the evaluation.}
    \label{fig:Scene_description_LC}
    \vspace{-1em}
\end{figure}

We consider the following three usecases, each one represented by a dedicated driving scenario.
We provide a description of the vehicle's mission in each scenario, along with a brief explanation of the corresponding augmented scene model.
\begin{description}[leftmargin=!,labelwidth=0.5em]
    \item[1) Following a multi-lane road] The vehicle follows the center of the current lane within a multi-lane road network.
     It adjusts its velocity and distance to the preceding traffic within its lane.
    This scenario describes typical ACC following behavior, but extends it by the awareness of the current lane geometry.
    The topological and geometric lane-level route is provided beforehand by a route planning module.
    The augmented scene contains the lane geometry along the planned route within in a predefined distance horizon.
    The perceived traffic participants are assigned to the lane network based on the geometric intersection of their bounding boxes with the lane geometry (see \citet{Ulbrich_Context_Representation_2014} for a more detailed description).
    Hence, an object needs to be located with at least one of its corners within the lane to get assigned to it.  
    The behavior planning module determines that one preceding traffic participant that requires the largest deceleration of the automated vehicle according to the \emph{Intelligent Driver Model} approach (cf. \citet{Treiber_IDM_2000}). 
     
    From the perspective of perception, objects inside the vehicle's lane have to be perceived with positions and dimensions accurate enough for a proper lane assignment. 
    Furthermore, objects in neighboring lanes are required to be perceived accurately enough so as not to be falsely associated to the vehicle's lane, as this would result in a wrong ACC target assignment.
     
    \item[2) Evaluating automated lane changes] This scenario extends the previous one by evaluating possible automated lane changes in dense traffic.
    Surrounding traffic has to be perceived to estimate proper gaps on neighboring lanes. 
    Hence, identical requirements for the objects' size and position estimations hold in this scenario, but need to include the backward traffic as well.
    For simplification, we assume only the next-left and -right neighbor lanes to be relevant. 
    Furthermore, we limit the number of objects to consider to four per lane: Two in front and two behind the ego vehicle each.
    This corresponds to the augmented scene model for tactical lane change decision making presented by \citet{Ulbrich_POMDP_2013, Ulbrich_Situation_Assesment_2015} (see \autoref{fig:Scene_description_LC}).
    Six regions of interests are defined, based on the lane geometries and their relative positions to the automated vehicle.
    The algorithms for object-to-lane assignment are identical to the previously described approach.
    This augmented scene model can be understood as a generalization of the ACC target selection subtask explained before.

    \item[3) Left turn through oncoming traffic]
    The vehicle has to perform an unprotected left turn through an intersection. 
    Thus, it has to ensure that no oncoming traffic will interfere with its turning maneuver.
    This scenario puts emphasizes on the detection range and the track setup latency, as the perception system needs to detect oncoming traffic at large distances and with high relative velocities.
    For sake of simplicity, we reduce this scenario to the estimation of existence of traffic participants located on the oncoming lanes and in the front region of the ego vehicle.
    The automated vehicle stops next to the intersecting area.
    The augmented scene model for this scenario thus only regards the geometry of these lanes and checks them for any valid association with detected traffic participants.
\end{description}

\subsection{Definition of datasets and evaluation criteria}
\label{ssec:Evaluation_Scenario_based}

Some popular datasets like \emph{KITTI} \citep{Geiger_KITTI_2012}, or the recent \emph{nuScenes} Dataset \citep{Caesar_NuScenes_2019}, already provide LiDAR pointclouds along with annotated ground-truth object information.
Despite the fact that using such a publicly available dataset would have allowed the comparison with other approaches, both expose some drawbacks regarding their usability for evaluation of the presented perception system (i.\,e., unlabeled regions or ranges \citep{Geiger_KITTI_2012}, no per-scan annotated data \citep{Caesar_NuScenes_2019}).
Thus, we use a preliminary version of the TUBS Road User Dataset  
\citep{Plachetka_TUBS_Dataset_2018} for evaluating our approach.
This dataset contains labeled objects within the entire field of view of the sensor and for each LiDAR scan, recorded along the project route of the project \emph{Stadtpilot}.

Based on the aforementioned scenario definitions, a subset of three sequences was selected.
The sequences contain  multi-lane roads and an intersection scenario with dense traffic.
In total, the subset comprises \num{1137}~scans with \num{9803}~labeled detections. 
Please note that the sequences contain several unlabeled scans prior to the actual reference data, in order to allow the perception system to initialize properly.

As stated above, \emph{open-loop} evaluation approaches are not meaningful to determine the perception system's ability to represent the \emph{relevant} objects for specific driving function/scenario configurations.
However, the \emph{close-loop} approach allows us to define the relevant traffic participants as part of the augmented scene model.
For the following evaluation, we consider those objects to be relevant that were mapped to the respective augmented scene description.
To cover time-varying relevancies, this assignment is evaluated in each time frame.
Differences of the relevance assignment between the system's output and the objects of the reference data set are accounted for as \emph{false positives} in the following metrics.

Evaluation is done twofold for the object configuration and the state estimation accuracy.
For the former, we use the MOTA score of the CLEAR MOT metrics, as introduced by \citet{Bernardin_CLEAR_2008}.
The \emph{Multiple Object Tracking Accuracy} is a measure for the object configuration estimation.
It subsumes the estimation of the correct number of objects and their consistent identification over time.
For a sequence of given length, it is defined as
\begin{align} \label{eq:Metric_MOTA}
\text{MOTA} &= 1 - \frac{\sum_{n} \text{FP}_n + \sum_{n} \text{FN}_n + \sum_{n} \text{IDsw}_n}{{\sum_{n} \text{RRO}_n}}
\end{align}
with FP and FN denoting false positives and negatives, and IDsw and RRO referring to the number of identity switches and \emph{relevant} reference objects at time frame~$n$, respectively.
\\
In the scope of this paper, a valid association of detected objects with references is defined as a non-zero overlap of their oriented bounding boxes.
We do not require a minimal intersection ratio, but rely on the state estimation evaluation to reflect potentially large deviations.

For the state estimation evaluation, we perform a per-state error distribution analysis. 
Independent error scores are calculated for the~$x$ and~$y$ position as well as for orientation and dimension differences.
By this, we gain insight into different error modes.
Each object's dimensions are evaluated using the length and width of the bounding box under consideration of the dimensions' quality classes. 
Values are compared only if the tracker considers the respective dimension as being fully visible, as explained in \autoref{ssec:Tracking_Occlusion_reasoning}.

\subsection{Results and discussion}
The following paragraphs show the results of the implemented object tracking system.
The mapping of the presented scenarios to the selected sequences of the data set was chosen as follows:
The evaluation of the \emph{Following a multi-lane road} and the \emph{Evaluating automated lane change} scenarios is performed on all three sequences, as all of them are located on multi-lane roads. 
The third sequence also contains a large intersection.
Hence, the evaluation of the \emph{Left turn through oncoming traffic} scenario is performed on the third sequence only.
We split the evaluation of the first two scenarios into the object configuration and the state estimation results.
For the third scenario, we give the results of the mean detection range and track initialization latency.

\paragraph{Object configuration}
\autoref{tab:Evaluation_Ergebnisse_MOTA_ACC_FSW} summarizes the results of the ACC following and lane change scenarios.
For ACC following, the perception system is able to reach a MOTA score of~\num{1.0}, thus being able to represent and select the correct ACC target object at all time frames.
The results of the lane change function are less sophisticated, yet showing a competitive performance.
Especially in the third sequence, a large false negative rate is accomplished.
This results from the second-order objects of the lane change relevance selection (see degradation of the MOTA score between the $\text{LC}_\text{all}$ and $\text{LC}_1$ results). In that particular sequence, these objects suffer from heavy occlusion due to other traffic participants. 
Thus, the tracker is not able to initialize tracks for certain time ranges, reducing the MOTA score.
The overall small numbers of object identity switches indicate that detected objects remain represented without any track losses or confusions. 

\begin{table}[ht]
    
    \caption{Achieved overall and per-sequence MOTA scores and error rates. 
        Results are given for both considered driving functions \emph{ACC} and \emph{automated lane changes} ($\text{LC}_\text{all/1}$).
        For sequence~\num{3}, also the scores considering only the first object per relevance region ($\text{LC}_1$) is provided.} 
    \label{tab:Evaluation_Ergebnisse_MOTA_ACC_FSW}
    \centering
    \sisetup{
        table-format=+1.2,
        table-number-alignment=center,
        table-auto-round
    }
    
    \begin{tabular}{ccSS[table-format=+1.1]S[table-format=+1.1]S[table-format=+2.0]}
        \toprule
        \multirow{2}{*}[-0.25em]{Seq.} & {\multirow[c]{2}{7em}[-0.25em]{\centering Driving function}}&  {\multirow{2}{*}[-0.25em]{MOTA}} &\multicolumn{2}{c}{Rates [\si{\percent}]} &  {\multirow{2}{*}[-0.25em]{IDsw}} \\
        \cmidrule(lr){4-5}
        & & &  {FPR} & {FNR} &  \\
        \cmidrule(lr){1-6}   
        \multirow{2}{*}{1}                 & $\text{LC}_\text{all}$ & 0.959   	 & 2.717391304347826 	 & 1.358695652173913 		 & 0    \\
        & ACC & 1.000     	 & 0.0 	 & 0.0 	  	 & 0    \\
        \addlinespace 
        \multirow{2}{*}{2}                & $\text{LC}_\text{all}$ & 0.9677  	 & 0.08733624454148471 	 & 2.1834061135371177 	  	 & 3 \\
        & ACC & 1.000    	 & 0.0 	 & 0.0 		 & 0 \\
        \addlinespace
        \multirow{3}{*}{3}      & $\text{LC}_\text{all}$ & 0.9036  	 & 0.472972972972973 	 & 8.64864864864865 	 	 & 7 \\
        
        & ACC & 1.000   	 & 0.0 	 & 0.0 	 	 & 0 \\
        & \textit{$\text{LC}_1^*$}      & 0.99   &0.4 & 0.5 &  2 \\      
        \cmidrule(lr){1-6}                      
        \multirow{2}{*}{overall}                     & $\text{LC}_\text{all}$ & 0.9222  	 & 0.537544095414077 	 & 6.954476734419621  	 & 10 \\
        & ACC & 1.000   	 & 0.0 	                & 0.0 	  	 & 0 \\
        \bottomrule   
        \addlinespace
    \end{tabular}
    {
        \footnotesize
        MOTA: Multiple Object Tracking Accuracy;  FPR/FNR: false positive/negative rates; IDsw: Number of track identity switches
    }  
    \vspace{-1.5em}
\end{table}

\begin{figure}
    \centering
    \begin{tikzpicture}[every node/.style={font=\footnotesize}]

\tikzstyle{background_histogram_style} = [fill=tubsBlueLight100, draw=tubsBlack, opacity=1.0]
\begin{groupplot}
	[
	group style={group size= 2 by 3, horizontal sep=0.25cm, vertical sep=0.95cm},
		width=4.75cm,
		height=3cm,
		ylabel = {norm. density},
		ybar,
		ymin = 0,
		xtick pos=left,
		grid=major,
		grid style={very thin, tubsGray40, dashed},	
		scaled y ticks=false,
		scaled x ticks=false,
		yticklabel style={
        	/pgf/number format/fixed,
        	/pgf/number format/fixed zerofill,
			/pgf/number format/set thousands separator={\,},
        	/pgf/number format/precision=1   
		},
		xticklabel style={
		    /pgf/number format/fixed,
        	/pgf/number format/fixed zerofill,
        	/pgf/number format/set thousands separator={\,},
        	/pgf/number format/precision=2, 
			yshift=0.35em        	
		},
		xlabel style={yshift=0.35em}
	]
	\nextgroupplot[
		xmin=-1.0, xmax=1.0, 
		ymax=4.0,
		xtick distance=0.75,
		bar width=3pt,
		xlabel = {$\Delta x$ [\si{\meter}]},
		yticklabel style={
        	/pgf/number format/fixed,
        	/pgf/number format/fixed zerofill,
			/pgf/number format/set thousands separator={\,},
        	/pgf/number format/precision=1   
		},
	]	

		\pgfplotstableread{\currentimagedirectory/histogram_x_error_overall_numdata.csv}\tmp
		\pgfplotstablegetelem{0}{Data}\of{\tmp}
		\pgfmathsetmacro{\NumData}{\pgfplotsretval}
		
		\addplot[
			background_histogram_style, 	
		] table[y expr=\thisrow{value}, x expr=\thisrow{bin}, col sep=semicolon]{\currentimagedirectory/histogram_x_error_overall_processed.csv};
		
		\node[anchor=north east] at (rel axis cs:1,1) {$N=\num[round-mode=places, round-precision=0]{\NumData}$};

\nextgroupplot[
	xmin=-1.0, xmax=1.0, 
	ymax=4.0,
	xlabel = {$\Delta y$ [\si{\meter}]}, 
	ylabel={}, 
	ylabel near ticks, 
	yticklabel pos=right,
	bar width=3pt,	
	xtick distance=0.75,
	yticklabel style={
        	/pgf/number format/fixed,
        	/pgf/number format/fixed zerofill,
			/pgf/number format/set thousands separator={\,},
        	/pgf/number format/precision=1   
		},
	]

		\pgfplotstableread{\currentimagedirectory/histogram_y_error_overall_numdata.csv}\tmp
		\pgfplotstablegetelem{0}{Data}\of{\tmp}
		\pgfmathsetmacro{\NumData}{\pgfplotsretval}
		
		\addplot[
			background_histogram_style, 	
		] table[y expr=\thisrow{value}, x expr=\thisrow{bin}, col sep=semicolon]{\currentimagedirectory/histogram_y_error_overall_processed.csv};
		
		\node[anchor=north east] at (rel axis cs:1,1) {$N=\num[round-mode=places, round-precision=0]{\NumData}$};

\nextgroupplot[
	xmin=-1.5, xmax=1.5, 
	ymax=4, 
	bar width=2.5pt,
	xlabel = {$\Delta l$ [\si{\meter}]},	
		xticklabel style={
	    /pgf/number format/fixed,
       	/pgf/number format/fixed zerofill,
       	/pgf/number format/set thousands separator={\,},
       	/pgf/number format/precision=1            	
		},
   ]
   
		\pgfplotstableread{\currentimagedirectory/histogram_length_error_overall_numdata.csv}\tmp
		\pgfplotstablegetelem{0}{Data}\of{\tmp}
		\pgfmathsetmacro{\NumData}{\pgfplotsretval}
	
	\addplot[
			background_histogram_style, 	
		] table[y expr=\thisrow{value}, x expr=\thisrow{bin}, col sep=semicolon]{\currentimagedirectory/histogram_length_error_overall_processed.csv};
		
		\node[anchor=north east] at (rel axis cs:1,1) {$N=\num[round-mode=places, round-precision=0]{\NumData}$};

\nextgroupplot[
	xmin=-1.5, xmax=1.5,
	ymax=4, 	
	bar width=2.5pt,
	xlabel = {$\Delta w$ [\si{\meter}]}, 
	ylabel={}, ylabel near ticks, yticklabel pos=right, 
		xticklabel style={
	    /pgf/number format/fixed,
       	/pgf/number format/fixed zerofill,
       	/pgf/number format/set thousands separator={\,},
       	/pgf/number format/precision=1            	
		},
	]
	
		\pgfplotstableread{\currentimagedirectory/histogram_width_error_overall_numdata.csv}\tmp
		\pgfplotstablegetelem{0}{Data}\of{\tmp}
		\pgfmathsetmacro{\NumData}{\pgfplotsretval}
		
		\addplot[
			background_histogram_style, 	
		] table[y expr=\thisrow{value}, x expr=\thisrow{bin}, col sep=semicolon]{\currentimagedirectory/histogram_width_error_overall_processed.csv};
		
		\node[anchor=north east] at (rel axis cs:1,1) {$N=\num[round-mode=places, round-precision=0]{\NumData}$};			

\nextgroupplot[
	xmin=-15, xmax=15, 
	ymax=0.4,
	width=8.25cm,
	bar width=2.9pt,
	xlabel = {$\Delta \Psi$ [\si{\LongDegree}] (orientation error)}, 
	xticklabel style={
	    /pgf/number format/fixed,
       	/pgf/number format/fixed zerofill,
       	/pgf/number format/set thousands separator={\,},
       	/pgf/number format/precision=0            	
		},	
	xtick distance=5,
	at = { ($ ( $ (group c1r2.south west) + (0,-60pt)$ )!0.5!(group c2r2.south east) $ ) }
	]
	
		\pgfplotstableread{\currentimagedirectory/histogram_yaw_error_overall_numdata.csv}\tmp
		\pgfplotstablegetelem{0}{Data}\of{\tmp}
		\pgfmathsetmacro{\NumData}{\pgfplotsretval}
		
		\addplot[
			background_histogram_style, 	
		] table[y expr=\thisrow{value}, x expr=\thisrow{bin}, col sep=semicolon]{\currentimagedirectory/histogram_yaw_error_overall_processed.csv};
		
		\node[anchor=north east] at (rel axis cs:1,1) {$N=\num[round-mode=places, round-precision=0]{\NumData}$};			
\end{groupplot}
	
\end{tikzpicture}
    \vspace*{-1em}
    \caption[]{Object state estimation results (error distributions) for different state dimensions. $N$ indicates the underlying sample size for each graph.}
    \label{fig:Evaluation_MOTP_Histograms}
    \vspace*{-1.5em}
\end{figure}
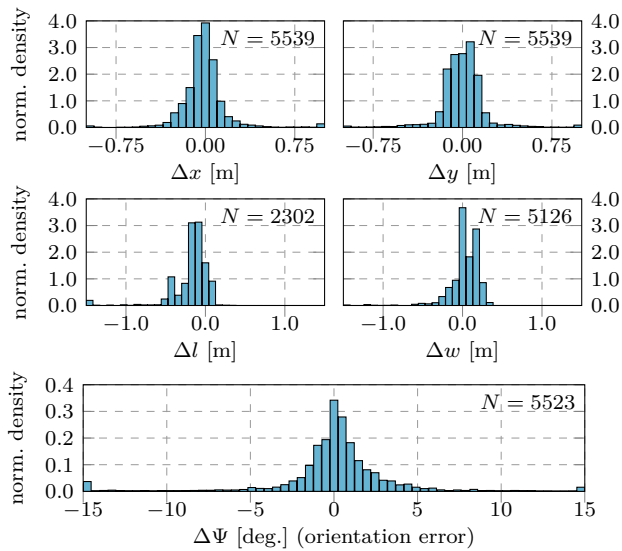

\paragraph{State estimation}
The object state estimation evaluation considers all true positive detections within the test sequences.
The resulting error distributions are shown in \autoref{fig:Evaluation_MOTP_Histograms}.
Position and orientation errors constitute a zero-symmetric distribution, with
average errors less than~\SI{20}{\centi\meter}/\ang{4} for position and orientation states, respectively.
Length and width errors show a skewed distribution with a tendency to underestimate the object dimensions. 
Outlier rejection filters and the sensor's susceptibility to total reflections at large inclination angles typically result in missing measurements, which lead to this effect in the end.

\paragraph{Detection latency and range}
We evaluate the detection latency and achievable detection range of the system in the intersection crossing scenario.
A total number of \num{16}~objects are driving on opposite lanes and are passing the ego vehicle.  
Their corresponding references appear at a mean distance of~\SI{65.5}{\meter}. 
The hypotheses generation is able to create box hypotheses after a mean time of \SI{390}{\milli\second}.
This delay arises from the fact that the ground surface estimation is unlikely to distinguish objects detected by only a single scan line from a ground surface candidate. 
The tracker is able to publish valid tracks after \SI{1.05}{\second}, resulting in a mean distance of \SI{53.3}{\meter}. 
Please note that a preceding car is occluding the direct view towards the oncoming traffic, which reduces the overall availability of raw measurements and thus reduces the perception system's performance in this particular scenario.

\subsection{Runtime performance}
Runtime performance is evaluated by re-processing a \num{27} minute test drive with a total number of \num{15745}~scans. 
All modules were implemented in C/C++ and use the \emph{Intel Threading Building Blocks} library\footnote{See \url{https://www.threadingbuildingblocks.org/}} for thread-level parallelization.
Evaluation was performed on an Intel i7-6700HQ-based platform (eight logical cores, \SI{3.2}{\GHz}).
To avoid unnecessary copy operations of pointcloud data, a copy-on-write scheme was applied to the pointcloud management infrastructure.
\\
On average, the results of the pointcloud preprocessing, the movable and the stationary environment model are available after \SI{39}{\milli\second}, \SI{52}{\milli\second} and \SI{60}{\milli\second}, respectively. 
Given an upper limit of \SI{100}{\milli\second} due to the sensor's rotation rate, the presented perception system is regarded as being real-time capable.

\section{Conclusion}
\label{sec:Conclusion}

In this paper, we presented a high-resolution LiDAR-based \ang{360} perception system for automated road vehicles in urban domains.
It is able to model 3D aspects of the environment and runs in real-time on consumer-grade computation hardware.
The architecture of the preprocessing stages is designed to deal with non-flat ground surfaces and protruding and overhanging structures.

We introduced extensions to our grid-based environment model regarding height filtering and an explicit free\-space representation.  
The model-based object tracking algorithm extends a widely known reference point approach by an explicit ambiguity notation and occlusion representation.
Evaluations using a novel scenario-based closed-loop methodology in combination with the TUBS Road User Dataset have shown impressive results  while still preserving real-time performance.

Within the evaluation, we assumed that the results on the chosen selection of real-world recordings and the definition of driving scenarios are able to reflect the general system performance.
Please note that, even though the authors intended to choose scenarios which represent typical traffic constellations and behavior for the goals of the project \emph{Stadtpilot} route, it is still ongoing research to provide proof for this assumption. 

Having a working perception system, the next steps will deal with the enhancement of different submodules.
In recent years, Neural Network-based approaches have shown promising results regarding pointcloud clustering and classification. 
We are keen on investigating the replacement of some parts of our perception pipeline with such approaches and compare the results.
Additionally, early fusion approaches, like presented by \citet{Held_Segmentation_2016} or \citet{Varga_Perception_Semantic_Pointcloud_Labeling_2017}, can be integrated to increase the performance of submodules throughout the perception system.

\section*{Acknowledgments}
The authors would like to thank the entire \emph{Stadtpilot} team, consisting of numerous PhD candidates, technical staff and lots of students, for their valuable work and support throughout the last years.



\renewcommand*{\bibfont}{\footnotesize} 
\printbibliography 

    \vskip-2.5em
\begin{IEEEbiography}[{\includegraphics[width=1in,clip,keepaspectratio]{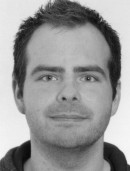}}]{Jens Rieken}
worked as a research assistant at the Institute of Control Engineering at TU Braunschweig since 2012 and is currently pursuing his PhD. He holds a Master of Science Degree in \emph{Electrical Engineering} from TU Braunschweig. His main research topics are environment perception and scene understanding for urban scenarios, as well as designing algorithms for real-time pointcloud processing.
\end{IEEEbiography}
    \vskip-2.5em
\begin{IEEEbiography}[{ \includegraphics[width=1in,clip,keepaspectratio]{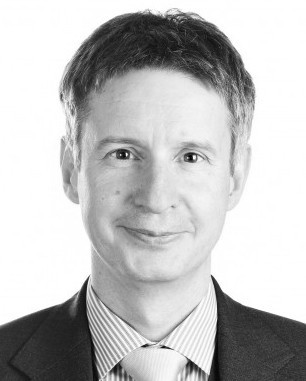}}]{Markus Maurer}
    holds the chair for Vehicle Electronics at TU Braunschweig since 2008. His main research interests include autonomous road vehicles, driver
    assistance systems, and automotive systems engineering. From 2000 to 2007 he was active in the development of driver assistance systems at Audi AG.
    \vspace*{-1em}
\end{IEEEbiography}

\newpage
%



\end{document}